\documentclass[10pt,journal,compsoc]{IEEEtran}

\ifCLASSOPTIONcompsoc
  \usepackage[nocompress]{cite}
\else
  \usepackage{cite}
\fi
\ifCLASSINFOpdf
\else
\fi
\usepackage{amsfonts, amssymb}
\usepackage{mathrsfs}
\usepackage{url}
\usepackage{graphicx}
\usepackage{amsmath}
\usepackage{booktabs}
\usepackage{multirow}
\usepackage{xcolor}
\usepackage{subfigure}
\usepackage{dsfont}
\usepackage{float}
\usepackage{bm}
\usepackage{algorithm}
\usepackage{algorithmic}
\begin{document}
%
\title{Learning Informative Representation for Fairness-aware Multivariate Time-series Forecasting: A Group-based Perspective}
%
%
%
%

\author{Hui~He,
        Qi~Zhang,
        Shoujin Wang,
        Kun~Yi,
        Zhendong~Niu,
        and~Longbing~Cao,~\IEEEmembership{Senior~Member,~IEEE}
\IEEEcompsocitemizethanks{\IEEEcompsocthanksitem Hui He is with the School of Medical Technology, Beijing Institute of Technology, Beijing 100081, China. \protect\\
E-mail: hehui617@bit.edu.cn
\IEEEcompsocthanksitem Qi Zhang is with the School of Computer Science and Technology, Beijing Institute of Technology, Beijing 100081, China, and the Department of Computer Science, Tongji University, Shanghai 201804, China.  \protect\\
E-mail: zhangqi\_cs@tongji.edu.cn
\IEEEcompsocthanksitem Shoujin Wang is with the Data Science Institute, University of Technology Sydney, Ultimo, NSW 2007, Australia.  \protect\\
E-mail: shoujin.wang@uts.edu.au
\IEEEcompsocthanksitem Kun Yi and Zhendong Niu are with the School of Computer Science and Technology, Beijing Institute of Technology, Beijing 100081, China. \protect\\
E-mail: \{yikun, zniu\}@bit.edu.cn
\IEEEcompsocthanksitem Longbing Cao is with the DataX Research Centre, School of Computing, Macquarie University, Sydney, NSW 2109, Australia.  \protect\\
E-mail: longbing.cao@mq.edu.au
}
\thanks{Manuscript received January 13, 2023; revised August 7, 2023; accepted
September 30, 2023. \\
(Corresponding authors: Qi Zhang and Zhendong Niu.)}}

%
%

\markboth{Journal of \LaTeX\ Class Files,~Vol.~14, No.~8, August~2015}%
{Shell \MakeLowercase{\textit{et al.}}: Bare Advanced Demo of IEEEtran.cls for IEEE Computer Society Journals}
%



\IEEEtitleabstractindextext{%
\begin{abstract}
Multivariate time series (MTS) forecasting penetrates various aspects of our economy and society, whose roles become increasingly recognized. However, often MTS forecasting is unfair, not only degrading their practical benefits but even incurring potential risk. Unfair MTS forecasting may be attributed to disparities relating to advantaged and disadvantaged variables, which has rarely been studied in the MTS forecasting. In this work, we formulate the MTS fairness modeling problem as learning informative representations attending to both advantaged and disadvantaged variables. Accordingly, we propose a novel framework, named \textit{FairFor}, for fairness-aware MTS forecasting, i.e., \textit{fair MTS forecasting}. \textit{FairFor} uses adversarial learning to generate both group-irrelevant and -relevant representations for downstream forecasting. \textit{FairFor} first adopts recurrent graph convolution to capture spatio-temporal variable correlations and to group variables by leveraging a spectral relaxation of the K-means objective. Then, it utilizes a novel filtering $\&$ fusion module to filter group-relevant information and generate group-irrelevant representations by orthogonality regularization. The group-irrelevant and -relevant representations form highly informative representations, facilitating to share the knowledge from advantaged variables to disadvantaged variables and guarantee the fairness of forecasting. Extensive experiments on four public datasets demonstrate the \textit{FairFor} effectiveness for fair forecasting and significant performance improvement.
\end{abstract}

}

\maketitle

\IEEEdisplaynontitleabstractindextext

%
\IEEEpeerreviewmaketitle

\ifCLASSOPTIONcompsoc
\IEEEraisesectionheading{\section{Introduction}\label{sec:introduction}}
\else
\section{Introduction}
\label{sec:introduction}
\fi

%
%
%
%

\begin{figure}[!t]
\centering
\includegraphics[width=1\linewidth]{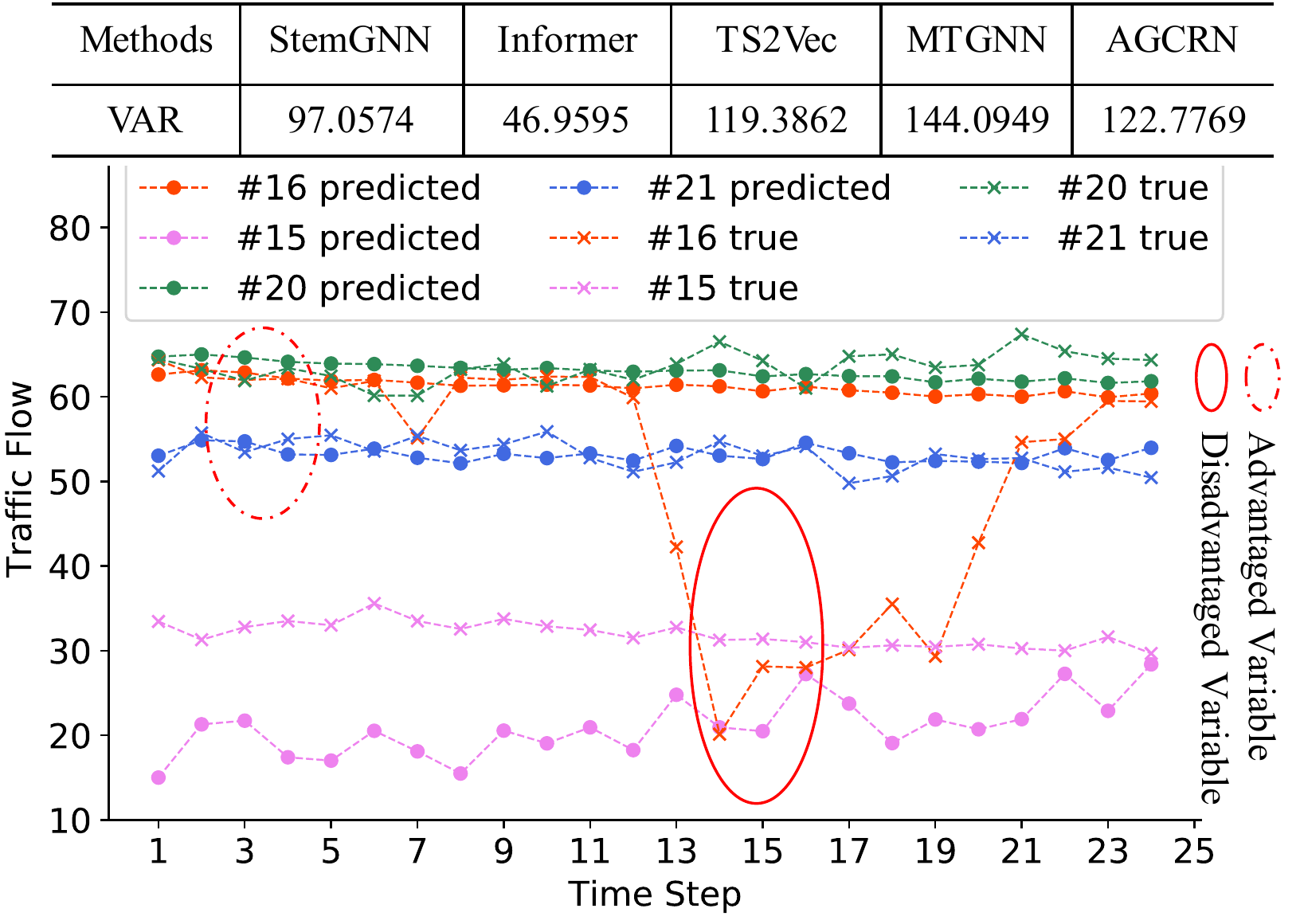}
\caption{An example of MTS forecasting unfairness: the table at the upper part shows the variance of the forecasting performance (MAE) over all variables derived from five advanced MTS forecasting methods on real-world traffic dataset; the curves below show the true traffic flow data and its corresponding prediction from StemGNN model on four sensors (variables) numbered 15, 16, 20 and 21. It is clear that StemGNN achieves desirable performance on those advantaged variables ($\#$20 and $\#$21 marked in dotted ellipse) but poor performance on those disadvantaged variables ($\#$15 and $\#$16 marked in solid ellipse).}

\label{figure1}
\end{figure}

\IEEEpubidadjcol

\IEEEPARstart{M}{ultivariate} time-series (MTS) forecasting penetrates our daily living, studying, working, and entertaining. It plays a critical role in increasingly wider real-world applications. Examples include COVID-19 forecasting \cite{Yangijcnn23,Yangdsaa23}, clinical risk forecasting~\cite{9794568}, financial forecasting~\cite{Xucao23, DBLP:conf/aaai/SawhneyAWDS21, 9815165}, climate forecasting~\cite{DBLP:conf/iclr/Shang0B21}, transport monitoring~\cite{DBLP:conf/nips/0001YL0020, DBLP:conf/icde/CirsteaYGKP22}, utility forecasting \cite{RenGLWYC22} and enterprise innovation analysis. 

However, an almost forgotten aspect in MTS forecasting is to evaluate the fairness of the modeling and results. Unfair MTS forecasting may result in unequal forecasting accuracy among variables and degrade the value of MTS forecasting or even bring about serious potential risk. 
For example, in clinical risk prediction, patients with rare diseases (e.g., Budd-Chiari syndrome) are easily misdiagnosed as having more common diseases (e.g., cirrhosis) since these diseases share similarities in pathological conditions (e.g., hepatomegaly), which easily overwhelm the distinct pathological conditions of rare diseases. In urban police deployment, the criminal activities across geographical neighborhood regions are easier to predict than isolated regions due to the shared socioeconomic factors. 
This may result in inherent variable disparity, e.g., patients with rare vs common diseases and isolated vs neighborhood regions, respectively, which may cause MTS forecasting models to generate unequal forecasting accuracy over different variables, i.e., well-performed (advantaged) variables and badly-performed (disadvantaged) variables in forecasting. This results in unfair results and overall performance, which may cause to distrust disadvantaged variables and potential practical failures or risks, 
e.g., fatal treatment plans or exploding crime rates.
Therefore, fair MTS forecasting is crucial, where one challenge is to balance both advantaged and disadvantaged variables for improving both the overall performance and specific performance on disadvantaged variables.

Various efforts aim to design competent MTS forecasting models with temporal dependencies (i.e., short- and long-term temporal patterns)~\cite{DBLP:conf/sigir/LaiCYL18, DBLP:journals/ml/ShihSL19, DBLP:conf/aaai/ZhouZPZLXZ21, DBLP:conf/iclr/LiuYLLLLD22}, spatial dependencies (i.e. inter-series correlations across multiple time series)~\cite{DBLP:conf/nips/0001YL0020, DBLP:conf/aaai/HeZBYN22, DBLP:conf/nips/CaoWDZZHTXBTZ20, DBLP:journals/pami/SpadonHBMRS22, DBLP:journals/pvldb/WuZGHYJ21}, modeling interpretability~\cite{DBLP:conf/ijcai/AssafS19, DBLP:conf/iclr/FortuinHLSR19, 9757812, DBLP:conf/icde/KieuYGJZHZ22}, and complicated nonstationary~\cite{DBLP:journals/corr/abs-2209-00654,DBLP:journals/tkde/YangYHM22}. Although achieving great success, these models focus on accuracy improvement~\cite{DBLP:journals/corr/abs-2010-03240} rather than variable disparity~\cite{DBLP:journals/csur/MehrabiMSLG21} to address performance unfairness among variables. As illustrated in Figure \ref{figure1}, five advanced MTS forecasting models have large accuracy variances (i.e., VAR) over different variables. StemGNN attends to certain variables and accordingly leads to obviously bad performance on variables $\#$15 and $\#$16 (disadvantaged variables). This illustrates the significant (widely existing) yet challenging (rarely considered) unfairness issue in MTS forecasting.



Although fair MTS forecasting has not been intensively studied, fairness modeling recently emerges in other learning tasks, such as classification~\cite{DBLP:conf/eccv/LinKJ22,DBLP:conf/aaai/ZhouMS21,DBLP:conf/aaai/ParkH0B21,9864304}, clustering~\cite{DBLP:conf/nips/NegahbaniC21,DBLP:conf/nips/EsmaeiliBSD21} and ranking~\cite{DBLP:conf/sigir/QiWWSWW0022, DBLP:conf/www/PatroBGGC20}, including by customizing fairness regularization~\cite{DBLP:conf/www/LiCFGZ21} and adversarial learning~\cite{DBLP:conf/www/WuCSHWW21,DBLP:conf/sigir/WuXZZ0ZL022}. These methods generally model fairness from two perspectives, i.e., individual and group perspectives. Some models individual fairness by a reasonable similarity metric based on a fairness graph~\cite{DBLP:conf/kdd/KangHMT20,DBLP:conf/kdd/DongKTL21}, weighted $\ell_p$-metrics~\cite{DBLP:conf/nips/NegahbaniC21,DBLP:conf/ijcai/YeomF20} or the Wasserstein distance~\cite{DBLP:conf/iclr/YurochkinBS20} at a fine granularity. Others aim to eliminate group unfairness on sensitive attributes (e.g., gender, age, or race of users). They learn to filter sensitive information via adversarial learning to obtain insensitive (group-irrelevant) representations~\cite{DBLP:conf/www/WuCSHWW21,DBLP:conf/sigir/WuXZZ0ZL022} or disentangle data representations into sensitive and insensitive representations by orthogonality regularization~\cite{DBLP:conf/sigir/QiWWSWW0022, DBLP:conf/www/PatroBGGC20}. However, individual fairness may be potentially harmful to overall forecasting performance and is costly in calculating the similarity measurement among all variables for high-dimensional MTS data~\cite{DBLP:conf/kdd/SongDLL22}. Group fairness generally depends on natural groups pre-defined by sensitive attributes, while such attributes are usually inaccessible in MTS data. This shows a challenge of fairness modeling in MTS scenarios.
Considering the inter-variable correlations in MTS data~\cite{DBLP:conf/aaai/HeZBYN22,RenGLWYC22}, correlated variables may have similar (advantaged/disadvantaged) performance and can be clustered into one group to handle. To this end, we deliberatively employ group fairness to achieve fair MTS forecasting, simplifying fairness modeling from the individual (variable) level to a group-wise manner and avoiding the high computational cost on individual variable fairness. We expect to share the knowledge between advantaged variables (groups) and disadvantaged variables (groups) to improve the performance of disadvantaged variables, guaranteeing performance fairness and overall improvement simultaneously. Accordingly, we address two essential challenges (CH): CH1, how can we adaptively learn variable grouping? CH2, how can we effectively learn group-relevant and -irrelevant representations? We propose a novel fair MTS forecasting model \textit{FairFor}, where two main modules: variable correlating $\&$ grouping and filtering $\&$ fusion,  address CH1 and CH2 respectively. In the variable correlating $\&$ grouping module, we first adopt recurrent graph convolution to capture spatio-temporal variable correlations and introduce clustering objectives to learn variable grouping. Inspired by~\cite{DBLP:conf/sigir/LiCXGZ21, DBLP:conf/www/WuCSHWW21}, we design a novel filtering $\&$ fusion module to generate group-irrelevant representations via adversarial learning. 


The contributions of our work are summarized below:
\begin{itemize}
\item We study a significant and common problem in MTS forecasting, i.e., the unfairness of forecasting performance. A novel fairness-aware MTS forecasting framework represents the first effort on fair MTS forecasting.
\item A novel variable correlating $\&$ grouping module learns spatio-temporal variable correlations by a recurrent graph convolutional network and adaptive variable grouping via the spectral relaxation of the K-means clustering objective.

\item A novel filtering $\&$ fusion module learns group-irrelevant representations by an orthogonality regularization in an adversarial learning framework.
\end{itemize}

Extensive experiments on four public datasets demonstrate the superior forecasting performance of \textit{FairFor} compared with the state-of-the-art models. We further verify the effectiveness and rationality of our proposed model in enhancing the fairness of MTS forecasting without performance loss.

\section{Related Work}
\subsection{Multivariate Time-series Forecasting}
Statistical methods for MTS forecasting, such as Gaussian process (GP)~\cite{DBLP:conf/nips/SalinasBCMG19}, vector auto-regressive model (VAR)~\cite{DBLP:journals/corr/abs-1910-11800} and autoregressive integrated moving average model (ARIMA)~\cite{DBLP:conf/aaai/ShiYCCYCYZ20}, all rely on powerful assumptions regarding a stationary process and can only learn linear relationship among different time steps within time series data. In contrast, deep learning based methods are immune to stationary assumptions and have an inherent efficiency in capturing non-linearity, complicated and hidden signals existing in many real-world time series collections. 
The first two deep learning based models created for MTS forecasting are LSTNet~\cite{DBLP:conf/sigir/LaiCYL18} and TPA-LSTM~\cite{DBLP:journals/ml/ShihSL19}. They marry convolutional neural network (CNN) with recurrent neural network (RNN) to capture short-range temporal dependencies and long-range temporal patterns respectively.
Informer~\cite{DBLP:conf/aaai/ZhouZPZLXZ21} embeds a sparse self-attention mechanism into Transformer-based architecture to enhance the latter's capacity in extracting the long-range dependencies and alleviate memory burden. Pyraformer~\cite{DBLP:conf/iclr/LiuYLLLLD22} explores the multi-resolution representation of time series via introducing a pyramidal attention module. Although these models are cutting-edge MTS forecasting algorithms, they only focus on exploiting temporal dependencies that deflate the ability when dealing with highly-correlated data. To further explicitly address the inter-series correlations among multiple variables, recent works extract the unidirected relations among variables and capture shared patterns via priori graph~\cite{DBLP:conf/ijcai/YuYZ18, DBLP:conf/iclr/LiYS018,DBLP:conf/ijcai/WuPLJZ19} or graph learning~\cite{DBLP:conf/nips/0001YL0020, DBLP:conf/nips/CaoWDZZHTXBTZ20, DBLP:conf/kdd/WuPL0CZ20, DBLP:journals/pvldb/CuiZCXDHZ21}. Similarly, CATN~\cite{DBLP:conf/aaai/HeZBYN22} introduces a tree (i.e., an ordered graph) to structure time-series variables with a clear hierarchy. More recent deep MTS models capture interactions across multiple MTSs \cite{RenGLWYC22,Yangijcnn23,Yangdsaa23} and high-dimensional dependencies \cite{Xucao23}. They rarely address the forecasting performance bias over different time series variables, while our work particularly focuses on fair MTS forecasting to avoid performance disparity on variables.  


\subsection{Fairness-aware Algorithms}
With the popularity of machine learning, the fairness issue has received broad attention as algorithms are vulnerable to data biases that render the decision skewed toward particular individuals. This data bias mostly relates to the issue of data imbalance (e.g., the historical user-item interactions between active users are much more than those between inactive users in recommendations) and then algorithms have been shown to amplify biases in the raw data to some extent.
Therefore, researchers have proposed many debiasing algorithms~\cite{DBLP:conf/kdd/KangHMT20,DBLP:conf/kdd/DongKTL21,DBLP:conf/iclr/YurochkinBS20,DBLP:conf/sigir/FuXGZHGXGSZM20,DBLP:conf/eccv/LinKJ22,DBLP:conf/aaai/ZhouMS21} to mitigate inequity for each specific domain. 
Specifically, some methods~\cite{DBLP:conf/kdd/KangHMT20,DBLP:conf/kdd/DongKTL21,DBLP:conf/iclr/YurochkinBS20} modeling individual fairness mostly define reasonable similarity metrics, but high-dimensional data make similarity measure between individuals very costly~\cite{DBLP:conf/kdd/SongDLL22}. Other methods fall into group fairness~\cite{DBLP:conf/eccv/LinKJ22,DBLP:conf/aaai/ZhouMS21} or a combination~\cite{DBLP:conf/sigir/FuXGZHGXGSZM20} of both group and individual fairness.
Lin et al.~\cite{DBLP:conf/eccv/LinKJ22} minimized the disproportionate impacts by calculating group-wise importance separately when pruning on different groups in the process of face classification model compression. 
Fu et al.~\cite{DBLP:conf/sigir/FuXGZHGXGSZM20} devised a fairness-constrained approach through heuristic re-ranking to mitigate the unfair recommendation issue where the user-item interactions of inactive users tend to be neglected and are easily overwhelmed by active users. 
In this work, we adopt group fairness to avoid high computational costs and achieve performance fairness and overall improvement simultaneously.


\subsection{Adversarial Learning for Fairness}
Next, a brief review is provided for adversarial learning algorithms applied in exploring fairness, which is the most relevant work to ours.
Some studies~\cite{DBLP:conf/sigir/LiCXGZ21,DBLP:conf/www/WuCSHWW21,DBLP:conf/sigir/WuXZZ0ZL022,DBLP:conf/www/PatroBGGC20,DBLP:conf/sigir/QiWWSWW0022,9864304} adversarially train models to discriminate sensitive attributes. 
For example, Li et al.~\cite{DBLP:conf/sigir/LiCXGZ21} learned a set of filters for erasing the sensitive attributes from the user representations and meanwhile use a set of classifiers as discriminators to predict sensitive attributes. Wu et al.~\cite{DBLP:conf/www/WuCSHWW21} also tried to obfuscate all sensitive attributes of users and items under a graph-based adversarial training process. Inspired by~\cite{DBLP:conf/acl/ColomboSNP22,DBLP:conf/aaai/ParkH0B21} disentangling data representation into orthogonal subspaces including sensitive attributes or not, 
Patro et al.~\cite{DBLP:conf/www/PatroBGGC20} simultaneously learned a bias-aware user representation and a bias-free user representation that only carries insensitive user information for fair news recommendation. 
Similarly, Qi et al.~\cite{DBLP:conf/sigir/QiWWSWW0022} designed an adversarial learning task to preclude encoding provider bias for provider-fair news representation and further render the provider-fair and provider-biased representations to be orthogonal by an orthogonal regularization term. 
However, these methods strongly depend on the pre-defined sensitive attributes that are unavailable in MTS analysis scenarios. In this study, we learn informative representations attending to each variable to promote fairness by leveraging a group-based adversarial learning framework. 

\begin{figure*}[!t]
\centering
\includegraphics[width=1\linewidth]{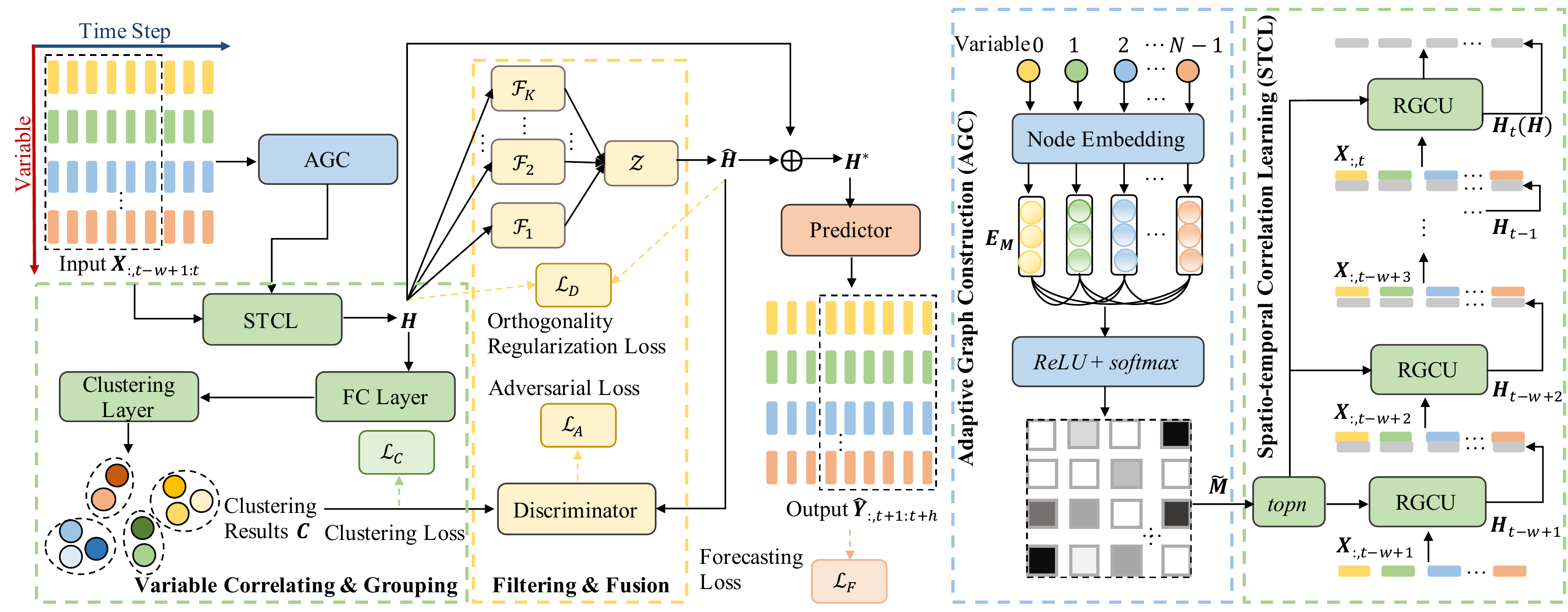}
\caption{The framework of our \textit{FairFor} network. AGC (the blue dotted line box) takes $\bm{X}_{:,t-w+1:t}$ as input to generate the learnable node embedding $\bm{E}_{\bm{M}}$ and adjacent matrix $\widetilde{\bm{M}}$. Then, the variable correlating $\&$ grouping (the green dotted line box) outputs the last hidden state $\bm{H}$ through $topn$ sparsization and multiple RGCUs and produces the clustering results $\bm{C}$. Next, the filtering $\&$ fusion (the yellow dotted line box) is applied to filter out the group-relevant information from $\bm{H}$ and form the group-irrelevant representation $\widehat{\bm{H}}$, which is further input to the discriminator $\mathcal{D}$ with $\bm{C}$. Lastly, the predictor integrates the group-relevant representation $\bm{H}$ and group-irrelevant representation $\widehat{\bm{H}}$ to form more informative representation $\bm{H}^{\star}$ for final prediction.}
\label{figure2}
\end{figure*}

\section{Methodology}
This section presents the problem formulation and then describes the overview of the proposed framework, named \textbf{fair}ness-aware multivariate time-series \textbf{for}ecasting (\textit{FairFor}) network, which are followed by the details of each module and the learning objectives.

\subsection{Problem Formulation}
We use $\bm{X}_{:,0:T-1}=\{\bm{X}_{:,0},\bm{X}_{:,1},...,\bm{X}_{:,t},...,\bm{X}_{:,T-1}\}\in\mathbb{R}^{T\times N}$ to denote $N$ univariate time series with $T$ time steps, where $\bm{X}_{:,t}=\{x_{0,t}, x_{1,t}, ..., x_{i,t}, ..., x_{N-1,t}\}^{\rm T} \in \mathbb{R}^{N \times 1}$ records the observed values of $N$ variables at time step $t$. Herein, $i \in \{0,1,...,N-1\}$ and $t \in \{0,1,...,T-1\}$ is the index of the variable and the time step respectively, and $\rm{T}$ denotes the transpose operation. The time interval between any two time steps is fixed. 

Under the sliding forecasting setting with a fixed window size of $w \in \mathbb{N}^+$ and a sliding step of 1, we have the input, i.e., the observed values of all the $N$ variables in $w$ successive steps till the $t^{th}$ time step, $\bm{X}_{:,t-w+1:t}=\{\bm{X}_{:,t-w+1},\bm{X}_{:,t-w+2},...,\bm{X}_{:,t}\}\in\mathbb{R}^{w\times N}$. We articulate the research problem of MTS forecasting on a graph $\bm{\mathcal{G}}=(\bm{\mathcal{V}}, \bm{\mathcal{E}}, \bm{M})$ to emphasize the spatio-temporal correlations simultaneously. The set of nodes $\bm{\mathcal{V}}$ denotes input series $\bm{X}_{:,t-w+1:t}$, where $|\bm{\mathcal{V}}|=N$ and each series/variable $\bm{X}_{i,:}$ corresponds to a node. $\bm{\mathcal{E}}$ represents the set of edges, and $\bm{M} \in \mathbb{R}^{N \times N}$ is defined as the adjacent matrix. In addition, we evaluate the forecasting unfairness with the variance in forecasting errors of different variables, where the larger the variance is, the more unfair the forecasts are. Accordingly, the target is to accurately forecast the future sequence of $h \in {\mathbb N}^+$ steps $\bm{Y}_{:,t+1:t+h}=\{\bm{Y}_{:,t+1},\bm{Y}_{:,t+2},...,\bm{Y}_{:,t+h}\}$ successive to the $t^{th}$ time step  based on graph $\bm{\mathcal{G}}$ through one forward procedure and guarantee a small forecasting error variance simultaneously.



\subsection{Framework Overview}
Figure \ref{figure2} illustrates an overview of \emph{FairFor} which consists of four modules: adaptive graph construction, variable correlating $\&$ grouping, filtering $\&$ fusion and predictor. To be specific, the adaptive graph construction is to learn an adjacent matrix (the implicit graph $\bm{\mathcal{G}}$) to capture the inter-series dependencies.
Then, based on the learned adjacent matrix, the variable correlating $\&$ grouping module learns a spatio-temporal representation for each time series with a recurrent graph convolutional network and then clusters all time series variables into multiple groups. Subsequently, the filtering $\&$ fusion module learns group-relevant variable representation and also group-irrelevant variable representation by filtering  the group-relevant information. Both the group-relevant representation and group-irrelevant representation are beneficial for the forecasting task. Therefore, the predictor integrates the two kinds of representations to form a more informative representation for producing the final forecast of each time series variable at one forward step.

\subsection{Adaptive Graph Construction}
Graph convolutional network (GCN) has been widely studied for accurate MTS forecasting due to its capability to capture the spatial correlations of time series data.  
Let $\bm{X} \in \mathbb{R}^{N \times w}$ (i.e., $w$-dimensional feature vector for each node) be the input time series matrix, $\bm{W}$ and $\bm{b}$ be the learnable weight matrix and bias respectively, then the layer-wise graph convolution operation of GCN can be well-approximated by the Chebyshev polynomial expansion form as: 
\begin{equation}\label{eqn1}
    \bm{\mathcal{G}}(\bm{X})=(\bm{I}+\bm{B}^{-\frac{1}{2}}\bm{M}\bm{B}^{-\frac{1}{2}})\bm{X}\bm{W}+\bm{b}
\end{equation}
where $\bm{I} \in \mathbb{R}^{N \times N}$, $\bm{B}$ is the degree matrix, and $\bm{\mathcal{G}}(\bm{X}) \in \mathbb{R}^{N \times q}$ is the output of the GCN layer. The GCN operation can be viewed from the perspective of a node (e.g., node $i$) as transforming the features of node $\bm{X}_i \in \mathbb{R}^{1 \times w}$ to $\bm{\mathcal{G}}(\bm{X}_i) \in \mathbb{R}^{1 \times q}$ with the shared $\bm{W}$ and $\bm{b}$.

The pre-defined adjacent matrix $\bm{M}$ embedded in the graph convolution operation is usually explicitly constructed by defining the similarity function of the dataset itself, or by defining the distance function according to the geographic distance on urban maps.
However, the fixed explicit graph structure is not always available or complete. The reason is that it is hard to manually capture latent relationships from substantial time-series data to construct the graph structure, especially for non-traffic datasets. Besides, the pre-defined graph is not directly related to the final fair forecasting task, which may lead to sizable biases. Inspired by ~\cite{DBLP:conf/nips/0001YL0020, DBLP:conf/nips/CaoWDZZHTXBTZ20, DBLP:conf/kdd/WuPL0CZ20, DBLP:journals/pvldb/CuiZCXDHZ21} precisely and automatically disclosing the implicit inter-series dependencies at each time step from MTSs without prior knowledge, 
in this paper, we propose an adaptive graph construction module to randomly initialize a learnable node embedding matrix $\bm{E}_{\bm{M}} \in \mathbb{R}^{N \times d}$ for all nodes. Herein, each row of $\bm{E}_{\bm{M}}$ represents the embedding of a node, and $d$ is the dimension of node embedding. Next, the inter-series dependencies between each pair of nodes can be inferred by multiplying $\bm{E}_{\bm{M}}$ and $\bm{E}_{\bm{M}}^{\rm T}$. Formally:
\begin{equation}\label{eqn2}
    \bm{B}^{-\frac{1}{2}}\bm{M}\bm{B}^{-\frac{1}{2}}=\delta(ReLU(\bm{E}_{\bm{M}} \cdot \bm{E}_{\bm{M}}^{\rm T}))
\end{equation}
where $\delta(\cdot)$ is the element-wise \emph{softmax} function which is used to normalize the adaptive adjacent matrix, $\rm{T}$ is the transpose operation. Here, to eliminate needless and repeated calculations during the iterative training process, we directly produce $\bm{B}^{-\frac{1}{2}}\bm{M}\bm{B}^{-\frac{1}{2}}$ in this case rather than producing $\bm{M}$ and computing a Laplacian matrix.  $\bm{E}_{\bm{M}}$ will be automatically updated throughout training to discover  hidden inter-series dependencies between various series and obtain  adaptive adjacent matrix $\bm{M}$ for graph convolutions. Therefore, the GCN enhanced by the adaptive graph construction can be formulated as:
\begin{equation}\label{eqn3}
    \bm{\mathcal{G}(X)} = (\bm{I}+\delta(ReLU(\bm{E}_{\bm{M}} \cdot \bm{E}_{\bm{M}}^{\rm T})))\bm{X}\bm{W}+\bm{b}
\end{equation}

\subsection{Variable Correlating $\&$ Grouping}
Accurate MTS forecasting relies on capturing two essential properties, i.e., temporal correlations over a sequence of time steps and spatial correlations over different time series variables. Accordingly, we adopt a recurrent graph convolutional network (RGCN) based on an implicit graph structure to capture the complex temporal and spatial correlations among a set of time series variables. We further group the learned spatio-temporal representation into several groups and produce the corresponding clustering results.



\textbf{Spatio-temporal Correlation Learning.} 
We introduce a recurrent graph convolutional unit (RGCU) via integrating a gated recurrent unit (GRU) with a GCN layer to learn the spatial and temporal inter-correlations between time series variables (see Figure \ref{figure2}). A GRU takes $\bm{X}_{:,t}$ as input and updates the hidden state from the previous state $\bm{H}_{t-1}$ to $\bm{H}_t$ by employing the reset gate and update gate to govern how much information from the history should be taken into account. By doing this, the GRU remembers historical hidden states that are relevant to future predictions and forgets those that are irrelevant. RGCU replaces the MLP layers in GRU with GCN and sends the embedded graph information to reset gate and update gate to update the hidden state $\bm{H}_t$ collectively. On the one hand, the spatial dependencies among different series, i.e., the inter-series dependencies, can be well captured by the implicit graph structure constructed based on the node embedding matrix $\bm{E}_{\bm{M}}$ in RGCU, where the implicit graph is encoded into graph nodes by GCN for time series variable interaction. On the other hand, the GRU structure can well capture the temporal dependencies over different time steps. As a result, the RGCU unit can well capture both temporal and spatial dependencies embedded in MTS data. Formally, we have  
\begin{equation}\label{eqn4}
    \widetilde{\bm{M}}=\delta(ReLU(\bm{E}_{\bm{M}} \cdot \bm{E}_{\bm{M}}^{\rm T}))
\end{equation}
\begin{equation}\label{eqn5}
    \widetilde{\bm{M}}_{i,\overline{\mathbf{idx}}}=0, s.t.\ \mathbf{idx}=topn(\widetilde{\bm{M}}_{i,:}), i \in \{0,1,...,N-1\}
\end{equation}
\begin{equation}\label{eqn6}
    \bm{r}_t=\sigma(\widetilde{\bm{M}}[\bm{X}_{:,t}||\bm{H}_{t-1}]\bm{W}_{\bm{r}}+\bm{b}_{\bm{r}})
\end{equation}
\begin{equation}\label{eqn7}
    \bm{u}_t=\sigma(\widetilde{\bm{M}}[\bm{X}_{:,t}||\bm{H}_{t-1}]\bm{W}_{\bm{u}}+\bm{b}_{\bm{u}})
\end{equation}
\begin{equation}\label{eqn8}
    \bm{c}_t=\theta(\widetilde{\bm{M}}[\bm{X}_{:,t}||\bm{r} \odot \bm{H}_{t-1}]\bm{W}_{\bm{c}}+\bm{b}_{\bm{c}})
\end{equation}
\begin{equation}\label{eqn9}
    \bm{H}_t=\bm{u} \odot \bm{H}_{t-1}+(1-\bm{u}) \odot \bm{c}_t
\end{equation}
where $\bm{X}_{:,t} \in \mathbb{R}^{N \times 1}$ and $\bm{H}_t \in \mathbb{R}^{N \times o}$ refer to the input and output at time step $t$, $o$ is the hidden dimension, $\bm{u}_t$ and $\bm{r}_t$ refer to the reset gate and update gate respectively, $\bm{c}_t$ is the memory state, $[\cdot||\cdot]$ denotes a concatenation operator, $\odot$ denotes an element-wise multiplication, $\sigma(\cdot)$ and $\theta(\cdot)$ are \emph{sigmoid} and \emph{tanh} activation functions. $\bm{E}_{\bm{M}}$, $\bm{W}_{\bm{u}}$, $\bm{W}_{\bm{r}}$, $\bm{W}_{\bm{c}}$, $\bm{b}_{\bm{u}}$, $\bm{b}_{\bm{r}}$ and $\bm{b}_{\bm{c}}$ are learnable parameters. Eq. (\ref{eqn5}) is a graph sparse strategy to make the adjacent matrix $\widetilde{\bm{M}}$ sparse and thus avoids introducing noise to the graph convolution. $topn(\cdot)$ selects the top-$n$ closest neighbors and accordingly returns their node indices $\mathbf{idx}$ for each node $i$. Subsequently, we retain the weights for selected nodes (associated with $\mathbf{idx}$) and set the weights of the unselected nodes (associated with $\overline{\mathbf{idx}}$) to zero. 
The last hidden state of RGCU denoted by $\bm{H}$ is sent to the variable grouping module and the filtering $\&$ fusion module respectively to generate clustering results and group-irrelevant representation respectively as the inputs to the discriminator $\mathcal{D}$.

\textbf{Variable Grouping.} Then, we group the learned hidden state $\bm{H}$ according to variable correlations and regard clustering/grouping results as a counterweight to the filtering $\&$ fusion module. Hence, we introduce a clustering layer consisting of a three-layer fully connected (FC) neural network with LeakyReLU as the activation and output the clustering results into discriminator $\mathcal{D}$ for adversarial learning. The learned hidden state $\bm{H}$ may not be suitable for forming cluster structures.
Hence, to stimulate to group variables in $\bm{H}$, we introduce the spectral relaxation of the K-means objective~\cite{DBLP:conf/aaai/MaCLC21,DBLP:conf/nips/MaZLC19} as a loss:
\begin{equation}\label{eqn10}
     \mathcal{L}_{C}=Tr(\bm{H}\bm{H}^{\rm T})-Tr(\bm{F}^{\rm T}\bm{H}\bm{H}^{\rm T}\bm{F}),\ s.t.\ \bm{F}^{{\rm T}}\bm{F}=\bm{I}
\end{equation}
where $\bm{H} \in \mathbb{R}^{N \times o}$, $\bm{F} \in \mathbb{R}^{N \times K}$ is the cluster indicator matrix and $K$ is the number of clusters. 
The optimization process of Eq. (\ref{eqn10}) requires iteratively updating $\bm{F}$ and $\bm{H}$ due to the dynamic learning of $\bm{H}$.
When $\bm{F}$ is fixed, updating $\bm{H}$ can follow the SGD of the clustering layer that boosts the representation to mine variable correlations. When $\bm{H}$ is fixed, $\bm{F}$ can be updated once by computing the $K$-truncated SVD of $\bm{H}$ after several epochs (e.g., 3 epochs) to prevent instability.  
Since the cluster-friendly representation $\bm{H}$ is not favorable for the forecasting process, it is difficult to jointly optimize $\bm{H}$ in the clustering and forecasting process. Hence, $\bm{H}$ is delivered to a FC layer before Eq. (\ref{eqn10}) to alleviate model instability, referring to Figure \ref{figure2}.

\subsection{Filtering $\&$ Fusion}
In our fairness-aware forecasting method, a key challenge is to learn the group-relevant (specific to each group) and group-irrelevant (shared by all groups) representations as illustrated in  \textit{Introduction}. Given a learning algorithm that learns spatio-temporal hidden state $\bm{H}$ to directly generate forecasts, we require the hidden state $\bm{H}$ to be independent of the learned group information $\bm{C}$ to achieve the forecasting fairness. Therefore, we design a filter layer with a series of filter functions $\{\mathcal{F}_1,\mathcal{F}_2,...,\mathcal{F}_K\}$, which are used to filter the group-relevant information in the hidden state $\bm{H}$ with $K$  groups. The filter function is represented as $\mathcal{F}:\mathbb{R}^{N \times o} \mapsto \mathbb{R}^{N \times o}$, and the representation $\mathcal{F}(\bm{H})$ preserves the features shared by all groups when specific features to per group are filtered out. We use the three FC layers followed by a batch normalization layer to represent each filter function $\mathcal{F}$. Finally, $K$ filtered representations are combined to generate the group-irrelevant representation:
\begin{equation}\label{eqn11}
    \widehat{\bm{H}} = \mathcal{Z}(\mathcal{F}_1(\bm{H}),\mathcal{F}_2(\bm{H}),...\mathcal{F}_K(\bm{H}))
\end{equation}
where $\mathcal{Z}$ is a fusion function. All $K$ filtered representations are fed into $\mathcal{Z}$ together and the group-irrelevant representation unrelated to all group information is output without the representation dimension altered, e.g., using the average of the $K$ filtered representations as $\mathcal{Z}$. Note that the number of filter functions in Eq. (\ref{eqn11}) does not necessarily equal  the number of groups $K$. To avoid introducing additional hyper-parameters, we set the number of filter functions to the number of groups $K$ here.

\textbf{Discriminator.}
To learn filter functions, we use the idea of adversarial learning to train a group discriminator. Specifically, we train two mappers $\mathcal{M}(\widehat{\bm{H}}):\mathbb{R}^{N \times o} \mapsto \mathbb{R}^{N \times o}$ and $\mathcal{M}(\bm{ C}):\mathbb{R}^{N \times K} \mapsto \mathbb{R}^{N \times o}$, which attempt to map the group-irrelevant representation and corresponding clustering results into the same space, more specifically, obtaining $\widehat{\bm{H}} \in \mathbb{R}^{N \times o}$ and $\bm{C} \in \mathbb{R}^{N \times o}$ respectively, and calculate their Euclidean distance. Similar to~\cite{DBLP:conf/nips/WuXDZ0H20}, we accordingly use the three FC layers with LeakyReLu as the activation function to represent $\mathcal{M}(\widehat{\bm{H}})$ and $\mathcal{M}(\bm{C})$ of discriminator $\mathcal{D}$. The aim of filter functions is to render its difficulty to infer variable correlations and group variables from the group-relevant representation $\bm{H}$, while that of discriminator $\mathcal{D}$ is to fail the filter functions.
The adversarial loss function is shown below:
\begin{equation}\label{eqn12}
     \mathcal{L}_{A}=\frac{1}{N}\sum\limits_{i=0}^{N-1}||\widehat{\bm{H}}_i-\bm{C}_i||_2^2
\end{equation}

Unfortunately, some group information may still be included in the group-irrelevant representation $\widehat{\bm{H}}$. Because the group-irrelevant representation $\widehat{\bm{H}}$ just needs to fool the discriminator $\mathcal{D}$, the filtering $\&$ fusion module does not necessarily completely filter the group information so that the discriminator $\mathcal{D}$ generally cannot perfectly assess the group information. To address this issue, we design an orthogonality regularization method~\cite{DBLP:conf/www/PatroBGGC20} to further purge the group-irrelevant representation. Specifically, the group-relevant representation $\bm{H}$ and group-irrelevant representation $\widehat{\bm{H}}$ are regularized by boosting them to be orthogonal to each other. The orthogonality regularization is calculated:
\begin{equation}\label{eqn13}
    \mathcal{L}_{D} = \frac{1}{N}\sum_{i=0}^{N-1}|\frac{\widehat{\bm{H}}_i \cdot \bm{ H}_i}{\|\widehat{\bm{H}}_i\| \cdot \|\bm{H}_i\|}|
\end{equation}
where $\widehat{\bm{H}}_i$ and $\bm{H}_i$ are the group-relevant and corresponding group-irrelevant representations of the $i^{th}$ variable.

\subsection{Prediction}
The group-relevant representation mainly contains information on group attributes, and the group-irrelevant representation mainly encodes group-free time-series information. Considering the information in both representations is relevant to the forecasting task, we integrate the group-relevant representation $\bm{H}$ and group-irrelevant representation $\widehat{\bm{H}}$ by an addition operation to form the informative representation $\bm{H}^{\star}$, i.e., $\bm{H}^{\star}=\bm{H}+\widehat{\bm{H}}$, which is then fed into a predictor. Then, the future time-series sequence is estimated by a 2-D convolutional layer at one forward step style rather than a step-by-step style:
\begin{equation}\label{eqn14}
    \widehat{\bm{Y}}=Conv2D(\bm{H}^{\star})
\end{equation}
where $Conv2D$ is a convolutional operation to directly map $\bm{H}^{\star}$ to the predictions for all horizons. 
Finally, the forecasting loss is denoted as:
\begin{equation}\label{eqn15}
     \mathcal{L}_{F}=\frac{1}{N}\sum_{i=0}^{N-1}||\bm{Y}_i-\widehat{\bm{Y}}_i||_2^2
\end{equation}

\begin{algorithm}[!t]
    \renewcommand{\algorithmicrequire}{\textbf{Require:}}
    \renewcommand{\algorithmicensure}{\textbf{}}
    \caption{Adversarial Training for \textit{FairFor}}
    \label{alg:1}
    \begin{algorithmic}[1]
        \REQUIRE
        Hidden state $\bm{H}$, group results $\bm{C}$, network $\mathcal{R}$, filter and fusion functions $\mathcal{F}, \mathcal{Z} \subseteq \mathcal{R}$, discriminator $\mathcal{D}$ and cluster number $K$
        \STATE Orthogonal initialize cluster indicator matrix $\bm{F}$
        \FOR{each training iteration $i$}
            \STATE $\widehat{\bm{H}} \leftarrow \mathcal{Z}(\{\mathcal{F}_k\}_{k \in [K]}(\bm{H}))$ by Eq. (\ref{eqn11}) \\
            \STATE $\bm{H}^{\star} \leftarrow \bm{H}+\widehat{\bm{H}}$ \\
            \STATE Optimize $\mathcal{L}$ w.r.t $\bm{H}$, $\widehat{\bm{H}}$, $\bm{H}^{\star}$, $\bm{C}$, $\mathcal{R}$ with $\mathcal{D}$ being fixed \\
            \IF{$i\%3==0$}
            \STATE Update $\bm{F}$ by computing $K$-truncated SVD of $\bm{H}$ \\
            \ENDIF
            \STATE $\mathcal{L}_A \leftarrow \mathbb{E}[||\widehat{\bm{H}}-\bm{C}||^2_2]$ by Eq. (\ref{eqn12})\\
            \STATE Optimize $\mathcal{L}_A$ w.r.t $\widehat{\bm{H}}$, $\bm{C}$, $\mathcal{D}$ with $\mathcal{R}$,  $\bm{H}$, $\bm{H}^{\star}$ fixed 
        \ENDFOR
    \end{algorithmic}  
\end{algorithm}

\subsection{Adversarial Training}
Adversarial learning techniques encourage the deep representation to be maximally informative to generate group-irrelevant representation, and meanwhile to be minimally discriminative in a group-relevant discriminator.
Therefore, adversarial learning has the potential to learn group-irrelevant representations and treat each variable fairly whether it is advantaged or disadvantaged. 
Note that we are not trying to purely render the prediction accuracy of advantaged variables and disadvantaged variables closer like common fairness-aware algorithms~\cite{DBLP:conf/www/PatroBGGC20,DBLP:conf/www/WuCSHWW21}. Instead, we try to learn the group-relevant and group-irrelevant representation by adversarial learning and then form informative representations focusing on both advantaged and disadvantaged variables for enhancing the performance on disadvantaged variables.
In the end, the \textit{FairFor} optimization involves playing a min-max game:
\begin{equation}\label{eqn16}
     \mathcal{L}=\arg\min_{\mathcal{R}}(\mathcal{L}_{F}+\mathcal{L}_{C}+\mathcal{L}_{D}+\\
     \arg\max_{\mathcal{D}}\lambda_a\mathcal{L}_A)
\end{equation}
where $\mathcal{R} = \textit{FairFor} - \mathcal{D}$ denotes the remaining part after removing discriminator $\mathcal{D}$ from \textit{FairFor}. The adversarial training algorithm is presented in Algorithm \ref{alg:1}. Our proposed \textit{FairFor} is carried out via alternately optimizing the subsequent processes. Concretely, we first feed input to the model to obtain $\mathcal{L}$, then fix the parameters in the discriminator $\mathcal{D}$, and optimize $\mathcal{R}$ by minimizing $\mathcal{L}$. Then, fixing the parameters  $\mathcal{R}$, $\mathcal{D}$ is optimized by minimizing $\mathcal{L}_A$.

\begin{table}[!t]
 \centering
 \caption{Dataset statistics.}
 \label{table1}
 \newcommand{\tabincell}[2]{\begin{tabular}{@{}#1@{}}#2\end{tabular}}
 \begin{tabular}{c|c|c|c|c}
  \toprule 
  Tasks & $\#$Time Step & $\#$Variable & Interval & Start Time\\
  \midrule 
  Traffic & 10,560 & 963 & 1hour & 1/1/2015 \\
  PeMSD7(M) & 11,232 & 228 & 5min & 7/1/2016 \\ 
  Solar-Energy & 52,560 & 137 & 10min & 1/1/2006 \\
  ECG5000 & 5,000 & 140 & $-$ & $-$ \\
  \bottomrule 
 \end{tabular}
\end{table}

\section{Experiments and Evaluation}
\subsection{Datasets}
To evaluate the models under different scales and application scenarios, we employ four real-world datasets (see dataset statistics in Table \ref{table1}) for extensive evaluation.

\textbf{PeMSD7(M) \footnote{ \url{https://dot.ca.gov/programs/traffic-operations/mpr/pems-source}}} records the traffic flow data of the detectors in California. It includes 228 variables and 11,232 time steps at a 5-minute interval;

\textbf{Solar-Energy \footnote{ \url{http://www.nrel.gov/grid/solar-power-data.html}}} is collected from National Renewable Energy Laboratory (NREL) and records the solar power production in 2006. It includes 137 variables and 52,560 time steps at a 10-minute interval; 

\textbf{Traffic \footnote{ \url{https://archive.ics.uci.edu/ml/datasets/PEMS-SF}}} is originally collected from the California Department of Transportation and describes the occupancy rate of different lanes in San Francisco highway. It contains 963 variables and 10,560 time steps with a 1-hour interval, where each observation is between 0 and 1;

\textbf{ECG5000 \footnote{ \url{http://www.timeseriesclassification.com/description.php?Dataset=ECG5000}}} records 5,000 heartbeats randomly selected from a 20-hour long ECG downloaded from Physionet. It contains 140 variables and 5,000 time steps.

\subsection{Experimental Setting and Metrics}
Following many classical time-series data split practices~\cite{DBLP:conf/nips/CaoWDZZHTXBTZ20,DBLP:conf/sigir/LaiCYL18}, we split the data into training, validation, and test parts with the ratio of 7:2:1. All data is normalized by the \emph{min-max} method by following~\cite{DBLP:conf/nips/CaoWDZZHTXBTZ20,DBLP:conf/nips/0001YL0020}. 
In our experiments, the FC layer, clustering layer, filters $\{\mathcal{F}_1,\mathcal{F}_2,...,\mathcal{F}_K\}$, discriminator $\mathcal{D}$ are all multi-layer FC networks with the number of layers set to 3. The predictor is a 2-D convolutional layer with the kennel size set to $(1,o)$ and the number of features $o$ in the hidden state $\bm{H}$ is fixed to 64. The cluster number $K$ in the filtering $\&$ fusion module is set to 6 for all datasets. 
We examine the values of trade-off parameter $\lambda_a$ in Eq. (\ref{eqn16}) in the range of $\{0,0.1,0.5.0.7,1\}$ and choose $\{\lambda_a=0.1\}$ for all datasets and the node embedding dimension $d$ in learnable node embedding matrix $\bm{E}_{\bm{M}}$ is set to 10. For network $\mathcal{R}$, we use an Adam optimizer with the learning rate of 3e-3. For discriminator $\mathcal{D}$, we use an Adam optimizer with the learning rate of 5e-2. Model parameters are turned on a Ubuntu 18.04.6 server with four NVIDIA GeForce 3090 GPU cards for 50 training epochs, through which we save the best-performing model based on values of accuracy metrics MAE/RMSE/MAPE/VAR on the validation set and reload it for the evaluation on the test set. The batch size is set as 64. 
Our codes are publicly available at https://github.com/huihevv/FairFor.

Four typical metrics, i.e., MAE, RMSE, MAPE, and VAR, are employed for MTS forecasting evaluation. Their formal definitions to evaluate all comparative methods are as follows: Mean Absolute Error $MAE=\frac{1}{N}\sum\limits_{i=0}^{N-1}|\bm{{\rm Y}}_i-\hat{\bm{{\rm Y}}}_i|$, Root Mean Squared Error $RMSE=\sqrt{\frac{1}{N}\sum\limits_{i=0}^{N-1}(\bm{{\rm Y}}_i-\hat{\bm{{\rm Y}}}_i)^2}$, Mean Absolute Percent Error $MAPE=\frac{1}{N}\sum\limits_{i=0}^{N-1}\frac{|\bm{{\rm Y}}_i-\hat{\bm{{\rm Y}}}_i|}{\bm{{\rm Y}}_i}\mathds{1}\{|\bm{{\rm Y}}_i|>0\}$ and Variance (the variance of MAE over each variable) $VAR=\frac{1}{N}\sum\limits_{i=0}^{N-1}[|\bm{{\rm Y}}_i-\hat{\bm{{\rm Y}}}_i|-\frac{1}{N}\sum\limits_{i=0}^{N-1}|\bm{{\rm Y}}_i-\hat{\bm{{\rm Y}}}_i|]^2$, where $N$ denotes the number of variables, $\bm{{\rm Y}}_i$ and $\hat{\bm{{\rm Y}}}_i$ denote the ground truth and prediction respectively.

\begin{table*}[!ht] 
 \centering
 \caption{The fairness performance of different methods evaluated on VAR under the setting of $w=12,h=12$, where the best results are highlighted in bold (smaller values indicate better fairness). 
}
\label{table2}
 \newcommand{\tabincell}[2]{\begin{tabular}{@{}#1@{}}#2\end{tabular}}
 \begin{tabular}{c|c|cc|c|cc|ccc|c}
  \toprule
  Dataset & Metric & LSTNet & TPA-LSTM & TS2Vec & Informer & Pyraformer & MTGNN & StemGNN & AGCRN & \textbf{\textit{FairFor}} \\
  \midrule 
  PeMSD7(M) & \multirow{4}{*}{VAR} & 23.6003 & 42.2336 & 28.4647 & 28.1218 & 48.4494 & 42.7716 & 26.5534 & \underline{22.7220} &\textbf{19.7870} \\
  \specialrule{0em}{1pt}{1pt}
  \cline{1-1} 
  \cline{3-11}
  \specialrule{0em}{1pt}{1pt}
  Solar-Energy & & 5.6880 & 10.1448 & 5.2330 & \underline{5.2096} & 10.6149 & 9.7419 & 5.7216 & 5.6543 & \textbf{4.7139} \\
  \specialrule{0em}{1pt}{1pt}
  \cline{1-1} 
  \cline{3-11}
  \specialrule{0em}{1pt}{1pt}
  Traffic & & 9.51e-4 & 8.07e-4 & 5.30e-4 & \underline{4.27e-4} & 7.63e-4 & 6.43e-4 & 6.71e-4 & 5.59e-4 & \textbf{4.11e-4} \\
  \specialrule{0em}{1pt}{1pt}
  \cline{1-1} 
  \cline{3-11}
  \specialrule{0em}{1pt}{1pt}
  ECG5000 & & 0.5668 & 0.2577 & 0.2360 & 0.2257 & 0.2618 & 0.2255 & \underline{0.1774} & 0.2467 & \textbf{0.1718} \\ 
 \bottomrule 
 \end{tabular}
\end{table*}

\begin{table*}[!ht]
 \caption{The prediction performance of different methods evaluated on four real-world datasets under the setting of $w=12,h=12$, where the best results are highlighted in bold (smaller values indicate better results). 
}
 \label{table3}
 \centering
 \newcommand{\tabincell}[2]{\begin{tabular}{@{}#1@{}}#2\end{tabular}}
 \begin{tabular}{c|c|cc|c|cc|ccc|c}
  \toprule
  Dataset & Metric & LSTNet & TPA-LSTM & TS2Vec & Informer & Pyraformer & MTGNN & StemGNN & AGCRN & \textbf{\textit{FairFor}} \\ 
  \midrule 
  \multirow{4}{*}{PeMSD7(M)} & MAE & 3.2004 & 4.7573 & 4.6001 & 3.9728 & 4.2925 & 4.4072 & 3.3064 & \underline{2.7983} & \textbf{2.7170} \\
  & RMSE & 5.6566 & 7.8292 & 7.0445 & 6.6654 & 8.1646 & 7.5602 & 5.7927 & \underline{5.4950} & \textbf{5.2009} \\
  & MAPE & 0.0870 & 0.1932 & 0.1117 & 0.0957 & 0.1067 & 0.1127 & 0.0817 & \underline{0.0694} & \textbf{0.0676} \\
  \midrule
  \multirow{3}{*}{Solar-Energy} & MAE & 1.3030 & 2.0399 & 1.3810 & 1.1426 & 2.2316 & 1.5043 & 1.1205 & \textbf{0.9090} & \underline{1.0476} \\ 
  & RMSE & 2.7057 & 3.1682 & 2.6721 & 2.5702 & 3.9558 & 2.7029 & 2.6720 & \textbf{2.5457} & \underline{2.5583} \\
  & MAPE & 3.4220 & 3.4175 & 3.3694 & 3.3826 & 3.4080 & 3.3727 & 3.4096 & \textbf{3.2905} & \underline{3.3539} \\
  \midrule
  \multirow{3}{*}{Traffic} & MAE & 0.0220 & 0.0158 & 0.0163 & \textbf{0.0095} & 0.0445 & 0.0117 & 0.0169 & 0.0123 & \underline{0.0103} \\ 
  & RMSE & 0.0374 & 0.0317 & 0.0282 & \textbf{0.0215} & 0.0523 & 0.0279 & 0.0310 & 0.0260 & \underline{0.0235} \\
  & MAPE & 0.7736 & 0.4924 & 0.5630 & \textbf{0.2482} & 2.2674 & 0.3429 & 0.4768 & 0.2676 & \underline{0.2581} \\
  \midrule
  \multirow{3}{*}{ECG5000} & MAE & 0.4978 & 0.4268 & 0.3634 & 0.3448 & \underline{0.3287} & 0.3493 & 0.3303 & 0.3534 & \textbf{0.3150}  \\ 
  & RMSE & 0.9022 & 0.6874 & 0.6067 & 0.5921 & 0.6110 & 0.5769 & \underline{0.5352} & 0.6096 & \textbf{0.5205} \\ 
  & MAPE & 0.9334 & 0.9561 & 0.9664 & 0.9481 & 1.0132 & 0.951 & \underline{0.9176} & 0.9854 & \textbf{0.8931} \\
  \bottomrule 
 \end{tabular}
\end{table*}

\subsection{Baselines}
Because the fairness problem in MTS forecasting has rarely been considered in previous studies, eight representative and SOTA MTS forecasting methods from different classes are deliberately chosen as baselines to compare with our proposed method. To be specific, two representative RNN-based methods, i.e., LSTNet and TPA-LSTM, and two Transformer-based methods, i.e., Informer and Pyraformer are chosen since they are good at capturing temporal patterns with different ranges such as short- and long-range. TS2Vec is a very popular universal time-series representation framework. Furthermore, MTGNN, StemGNN and AGCRN are based on graph neural networks, which are selected to justify the effectiveness of incorporating both intricate temporal and spatial dependencies among time-series data in our method.
Traditional methods such as VAR and GP are not compared since the latest deep neural network-based methods~\cite{DBLP:conf/kdd/WuPL0CZ20,DBLP:journals/pvldb/CuiZCXDHZ21,DBLP:conf/aaai/0002CHP22} have been verified to outperform these methods.

For a fair comparison, the length of time-series input (historical sequence), time-series output (future sequence), and hardware technical indicators for all baselines are identical. Hyper-parameters of each baseline are consistent with the corresponding paper settings. Some methods~\cite{DBLP:conf/sigir/LaiCYL18, DBLP:journals/ml/ShihSL19} were originally proposed for single-step output, then we carefully modify them into sequence output. More details about these methods are as follows:

\textbf{LSTNet}~\cite{DBLP:conf/sigir/LaiCYL18}: integrates RNNs and CNNs to capture the short- and long-term temporal patterns respectively; 

\textbf{TPA-LSTM}]~\cite{DBLP:journals/ml/ShihSL19}: embeds temporal pattern attention into RNNs to discover both relevant time series and time steps;


\textbf{TS2Vec}~\cite{DBLP:conf/aaai/YueWDYHTX22}: is a dilated CNN-based universal framework to capture multi-scale contextual information in MTSs;  

\textbf{Informer}~\cite{DBLP:conf/aaai/ZhouZPZLXZ21}: is a Transformer-based model with \emph{ProbSparse} self-attention mechanism and generative style decoder to predict long time series at one forward step; 

\textbf{Pyraformer}~\cite{DBLP:conf/iclr/LiuYLLLLD22}: is a Transformer-based model to simultaneously extract multiple ranges of temporal dependencies via the compact multi-resolution operation;

\textbf{MTGNN}~\cite{DBLP:conf/kdd/WuPL0CZ20}: combines graph learning, graph convolution and temporal convolution together to learn the spatial-temporal correlations without pre-defined graph structure;

\textbf{StemGNN}~\cite{DBLP:conf/nips/CaoWDZZHTXBTZ20}: uses a GCN-based spectral network that can capture inter-series dependencies and temporal correlations jointly in the spectral domain;

\textbf{AGCRN}~\cite{DBLP:conf/nips/0001YL0020}: employs GCNs embedded with an adaptive node-specific pattern learning module to capture fine-grained inter-series relationships and uses RNNs to capture temporal patterns.

\subsection{Experimental Results}

\subsubsection{Overall Results}
We compare \textit{FairFor} with the baseline models on both MTS forecasting and fairness performance. Table \ref{table2} and Table \ref{table3} show the fairness and forecasting performance of different methods respectively under the commonly-used setting of $w=12,h=12$ following~\cite{DBLP:conf/nips/0001YL0020,DBLP:conf/kdd/WuPL0CZ20}. 

\textbf{Fairness Improvement} In the fairness aspect, we find that our \textit{FairFor} obviously outperforms all baseline MTS forecasting methods on four datasets, specifically the VAR decrease over the best baseline is 12.92$\%$/9.52$\%$/3.75$\%$/3.16$\%$ on PeMSD7(M)/Solar-Energy/Traffic/ECG5000 respectively. This is consistent with previous analysis: 1) general forecasting models are vulnerable to variable disparity and prone to focus on certain variables (advantaged variables), leading to generating unequal performance over different variables; 2) \textit{FairFor} proposes to combine group-relevant and group-irrelevant representation together to form informative representation focusing on both advantaged and disadvantaged variables. Consequently, the performance of disadvantaged variables is improved by enriching the group-irrelevant representation and drawing support from the knowledge of advantaged variables.

\begin{figure*}[!t]
\centering
\subfigure[PeMSD7(M)]{
    \label{PeMSD7(M)_MAE}
    \begin{minipage}{0.236\linewidth}
        \includegraphics[width=1\linewidth,height=100pt]{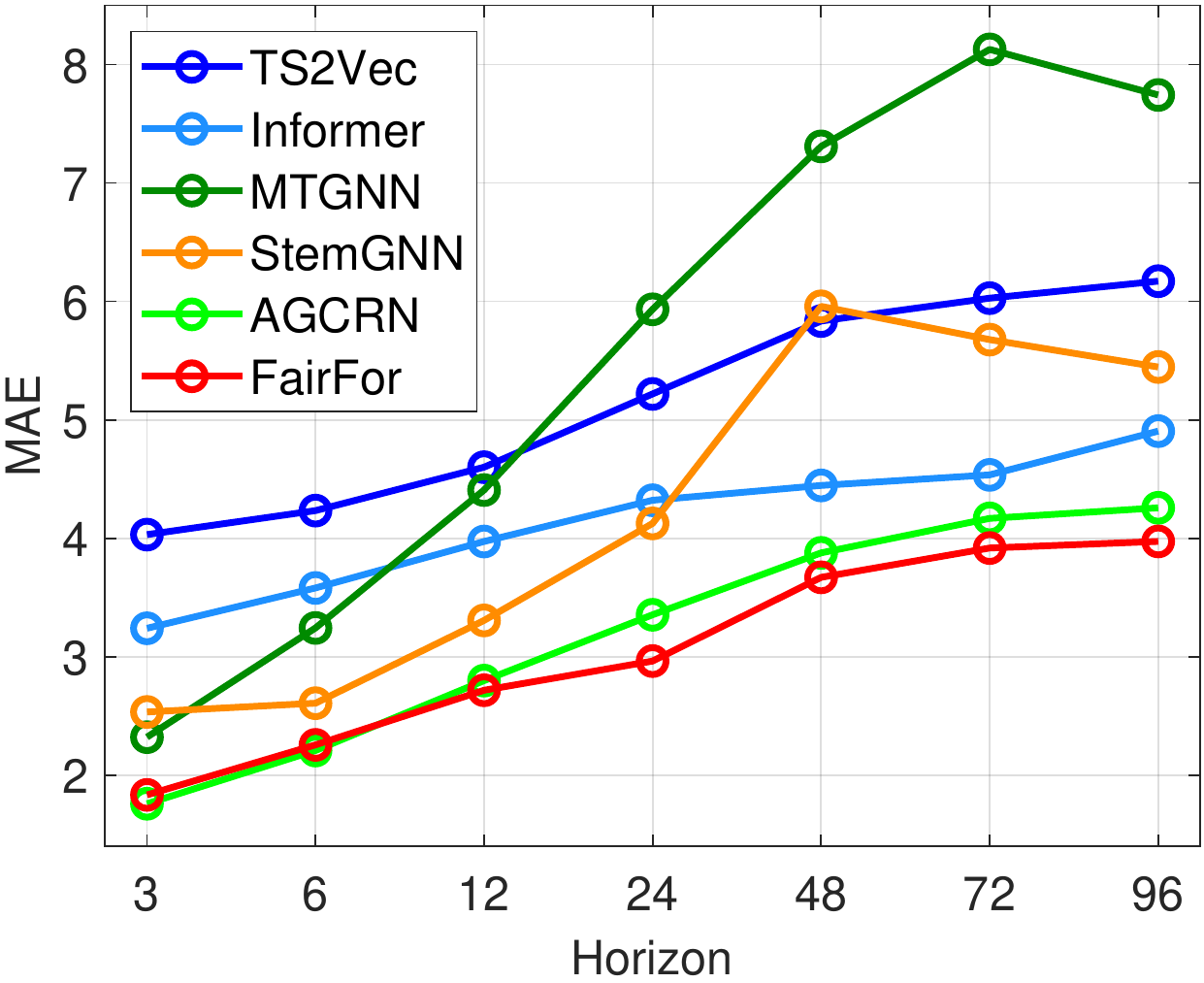}\\
        \vspace{0.3mm}
        \includegraphics[width=1\linewidth,height=100pt]{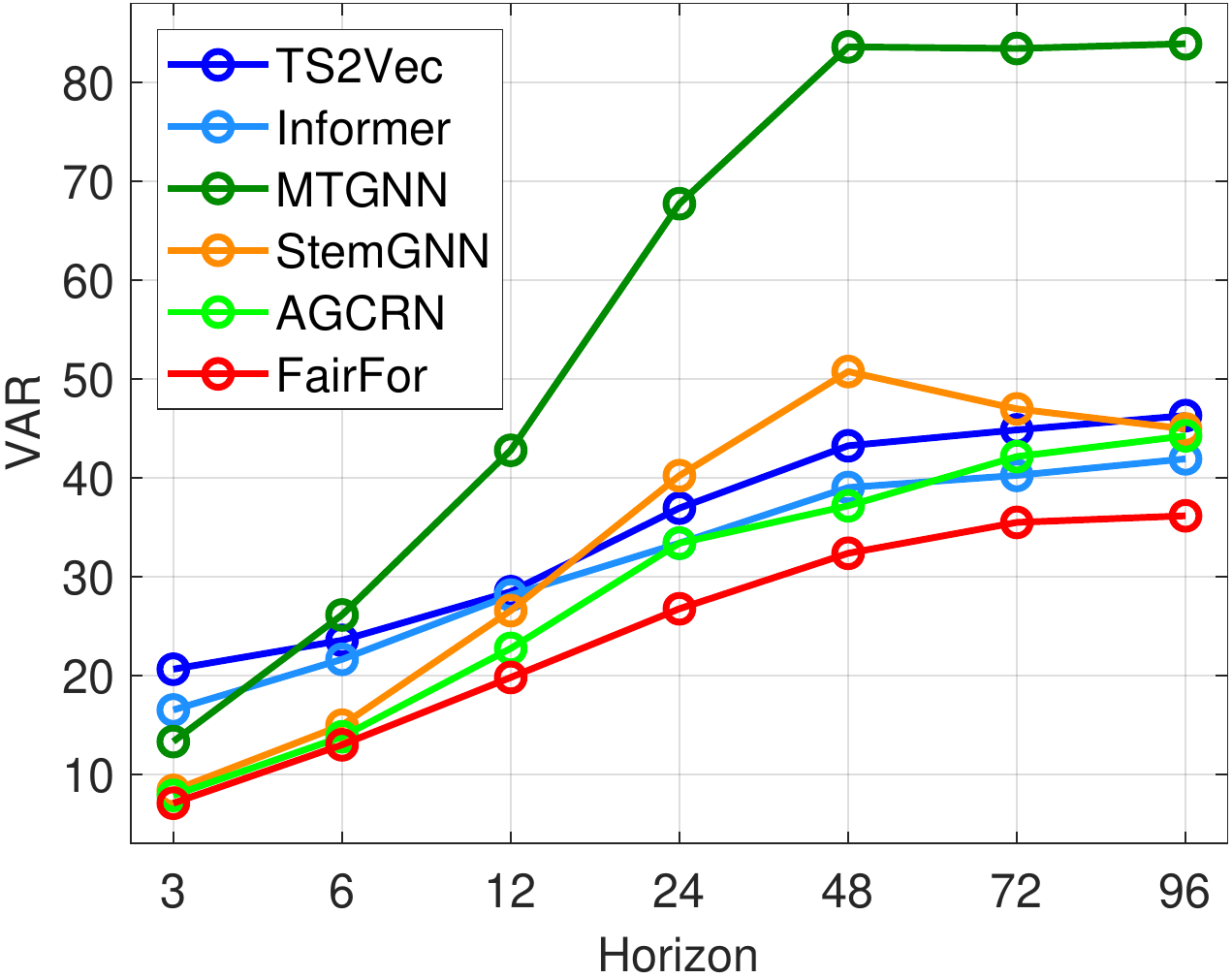}
        \vspace{0.3mm}
    \end{minipage}
    }
\subfigure[Solar-Energy]{
    \label{Solar-Energy_MAE}
    \begin{minipage}{0.236\linewidth}
        \includegraphics[width=1\linewidth,height=100pt]{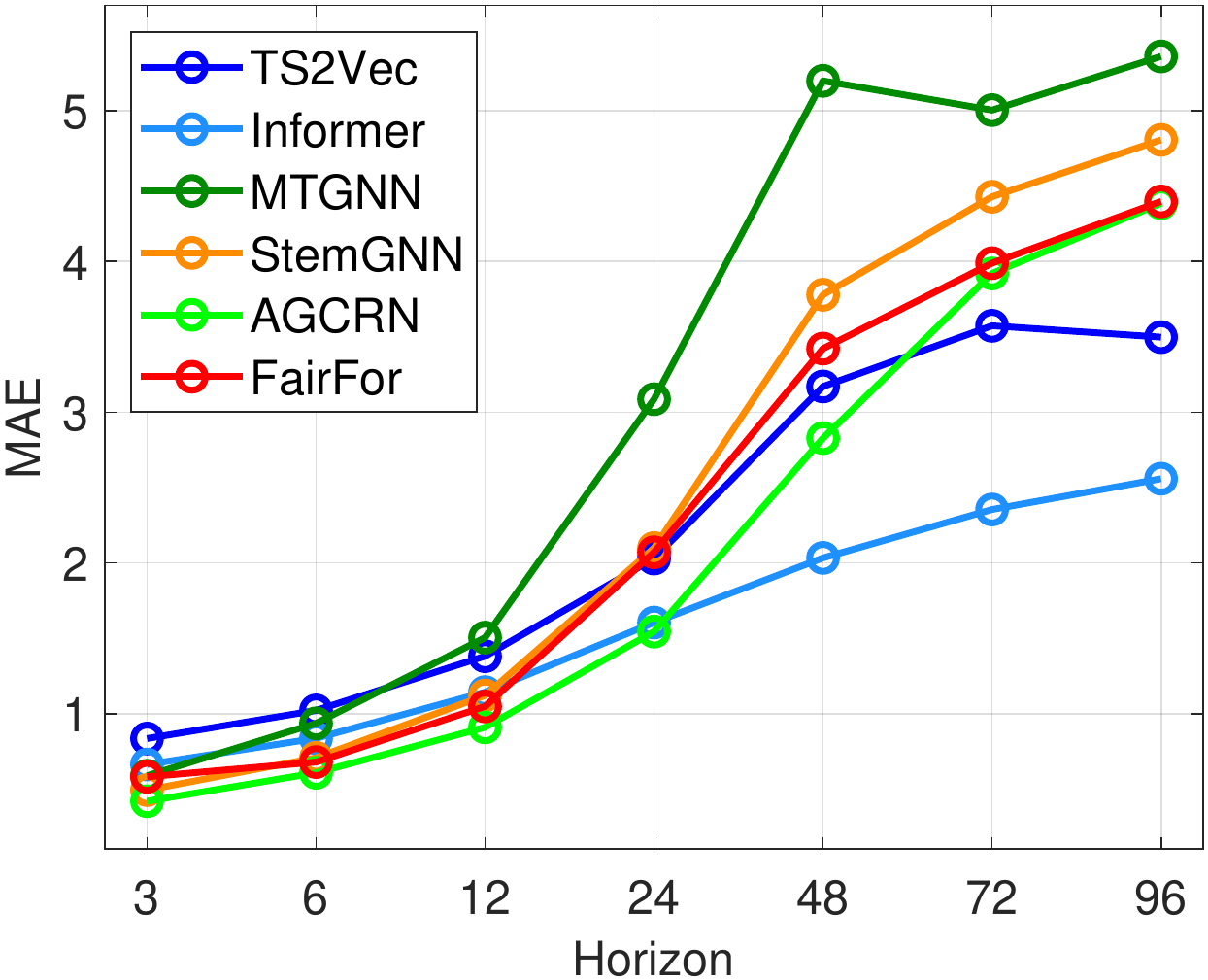}\\
        \vspace{0.3mm}
        \includegraphics[width=1\linewidth,height=100pt]{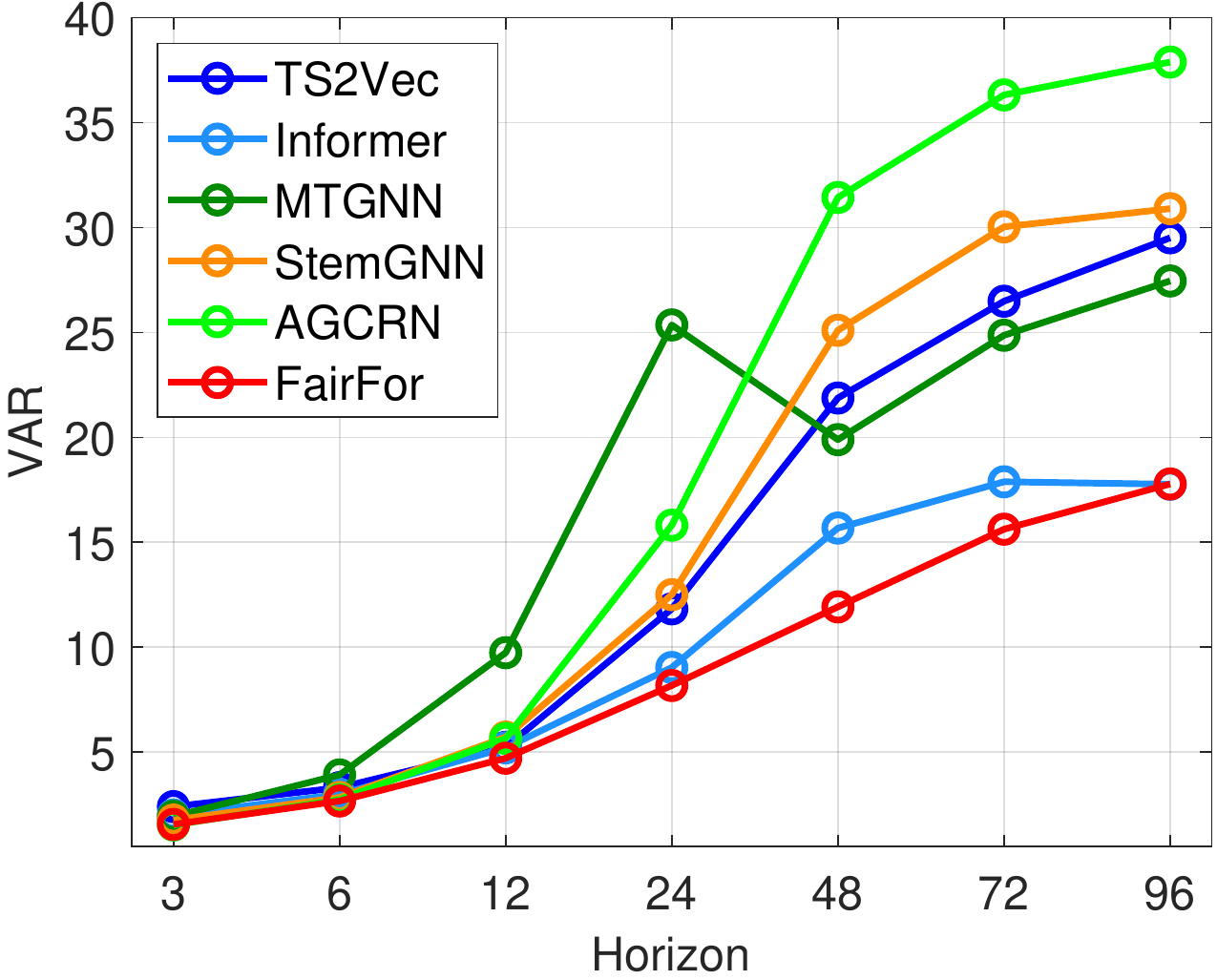}
        \vspace{0.3mm}
    \end{minipage}
    }
\subfigure[Traffic]{
    \label{Traffic_MAE}
    \begin{minipage}{0.236\linewidth}
        \includegraphics[width=1\linewidth,height=100pt]{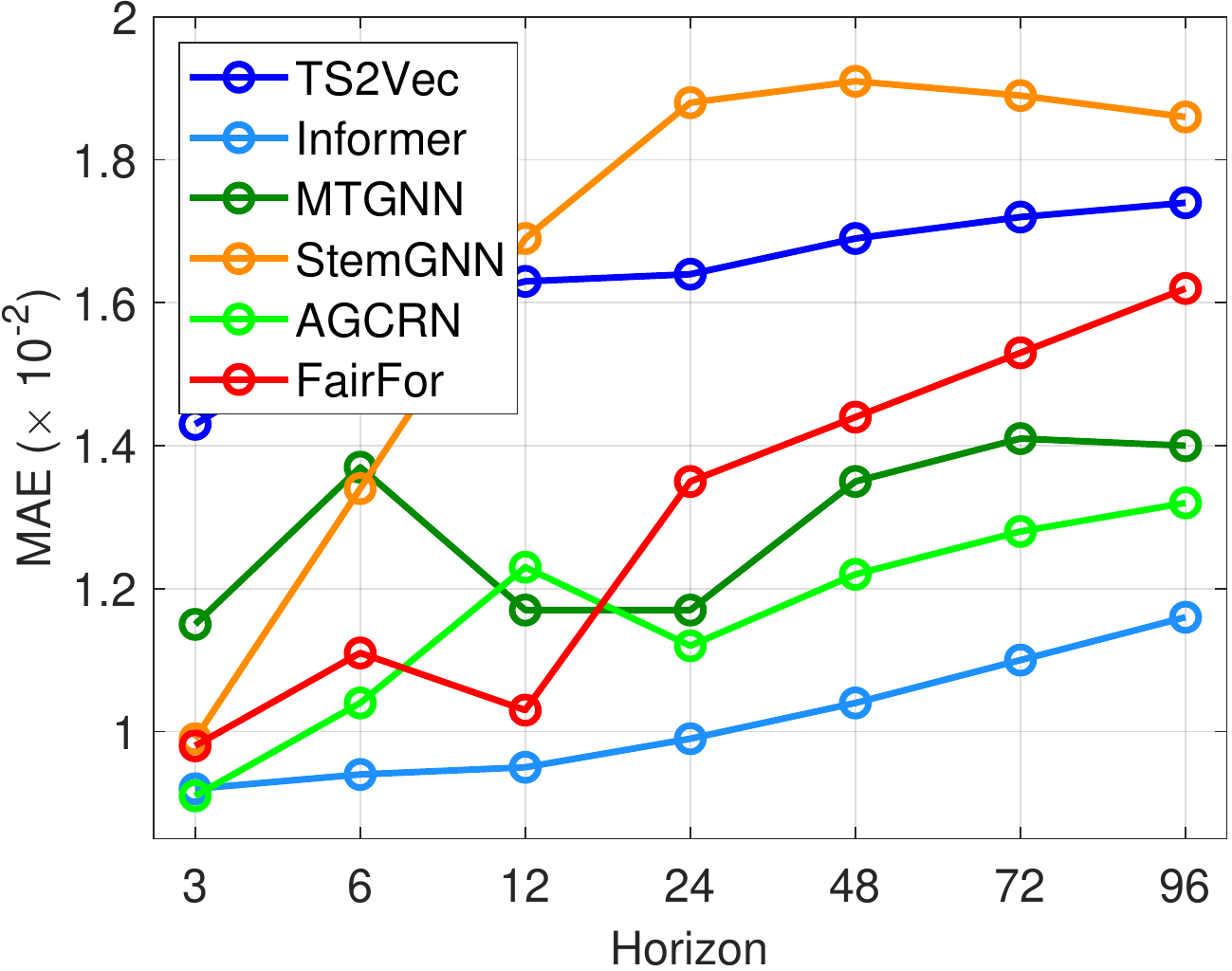}\\
        \vspace{0.3mm}
        \includegraphics[width=1\linewidth,height=100pt]{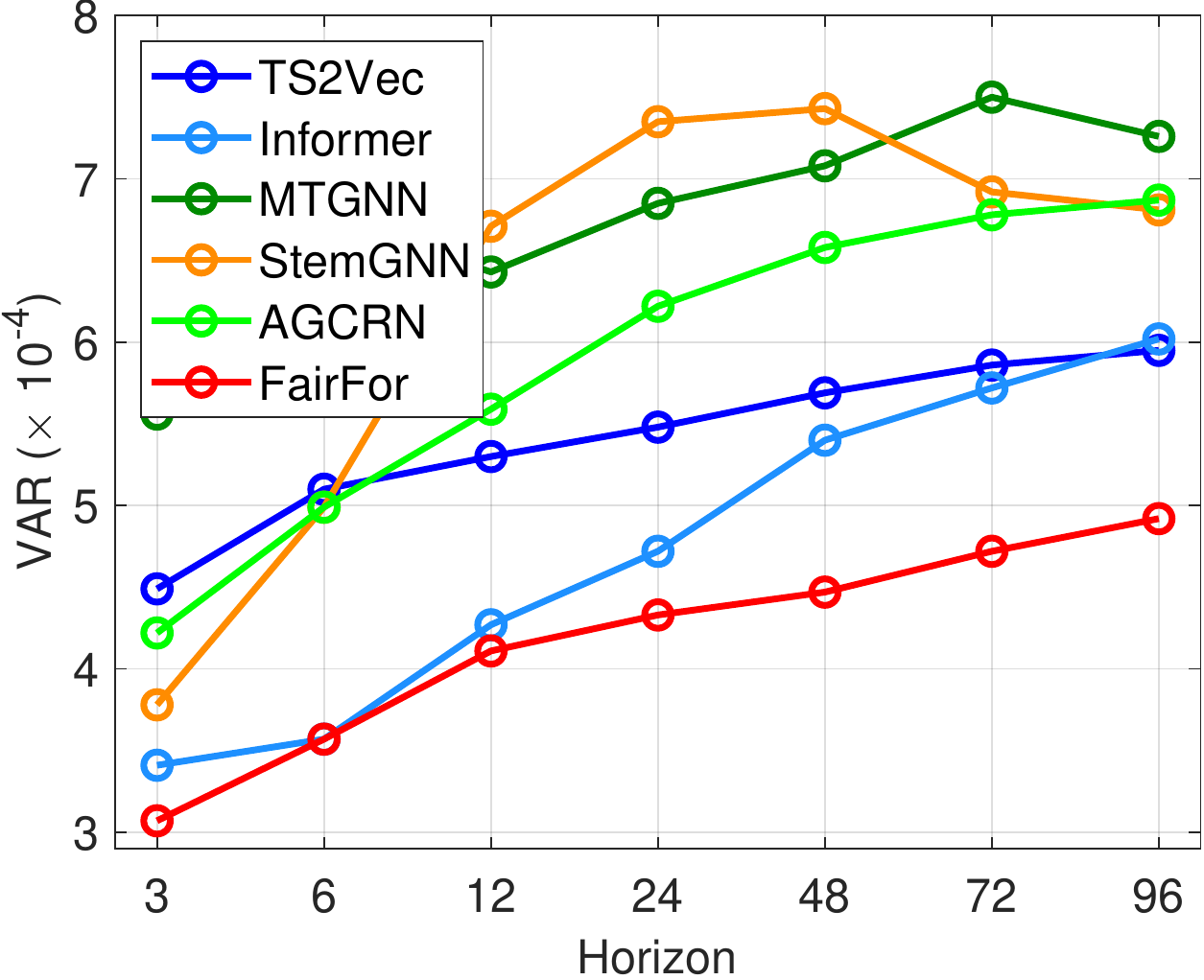}
        \vspace{0.3mm}
    \end{minipage}
    }
\subfigure[ECG5000]{
    \label{ECG5000_MAE}
    \begin{minipage}{0.236\linewidth}
        \includegraphics[width=1\linewidth,height=100pt]{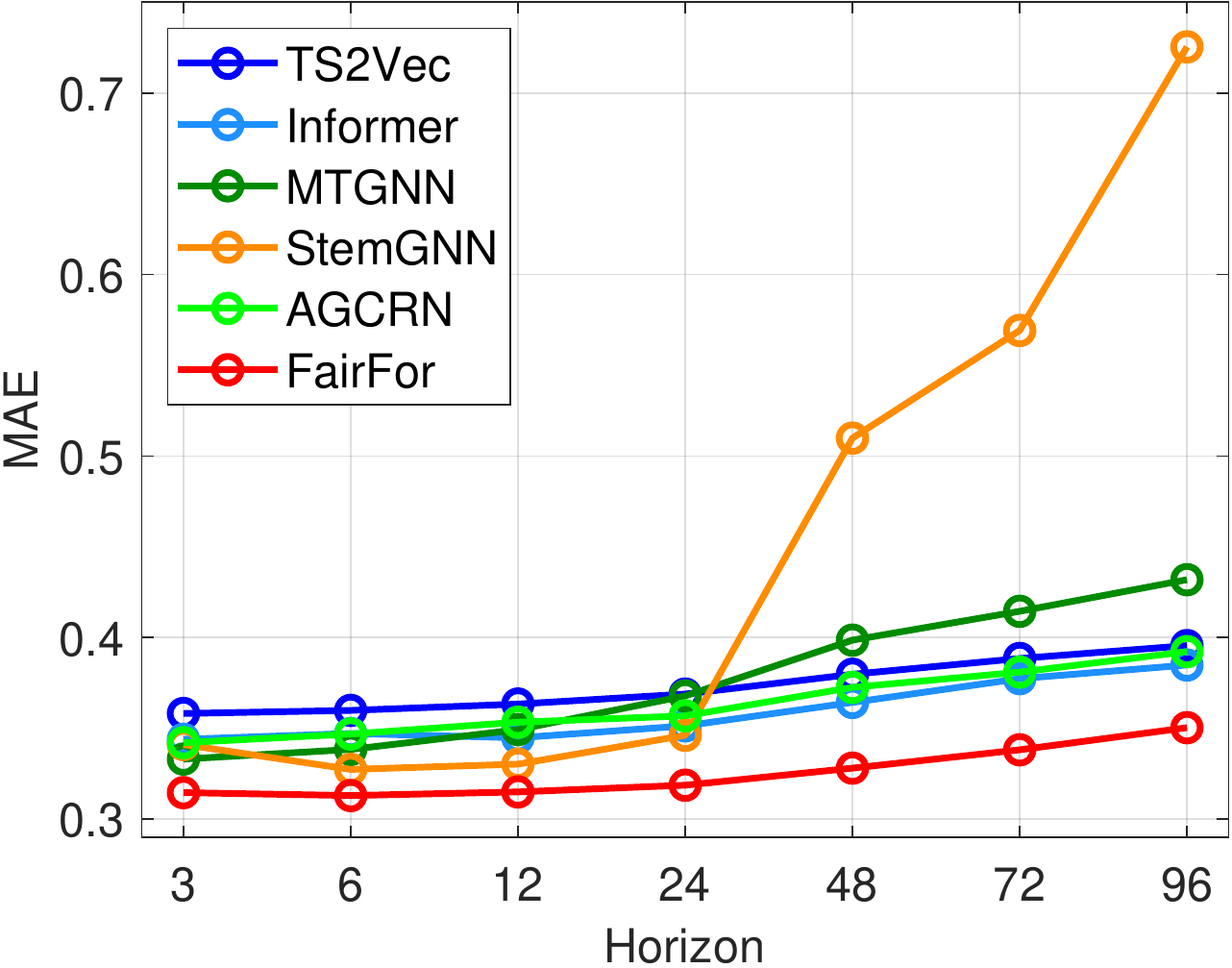}\\
        \vspace{0.3mm}
        \includegraphics[width=1\linewidth,height=100pt]{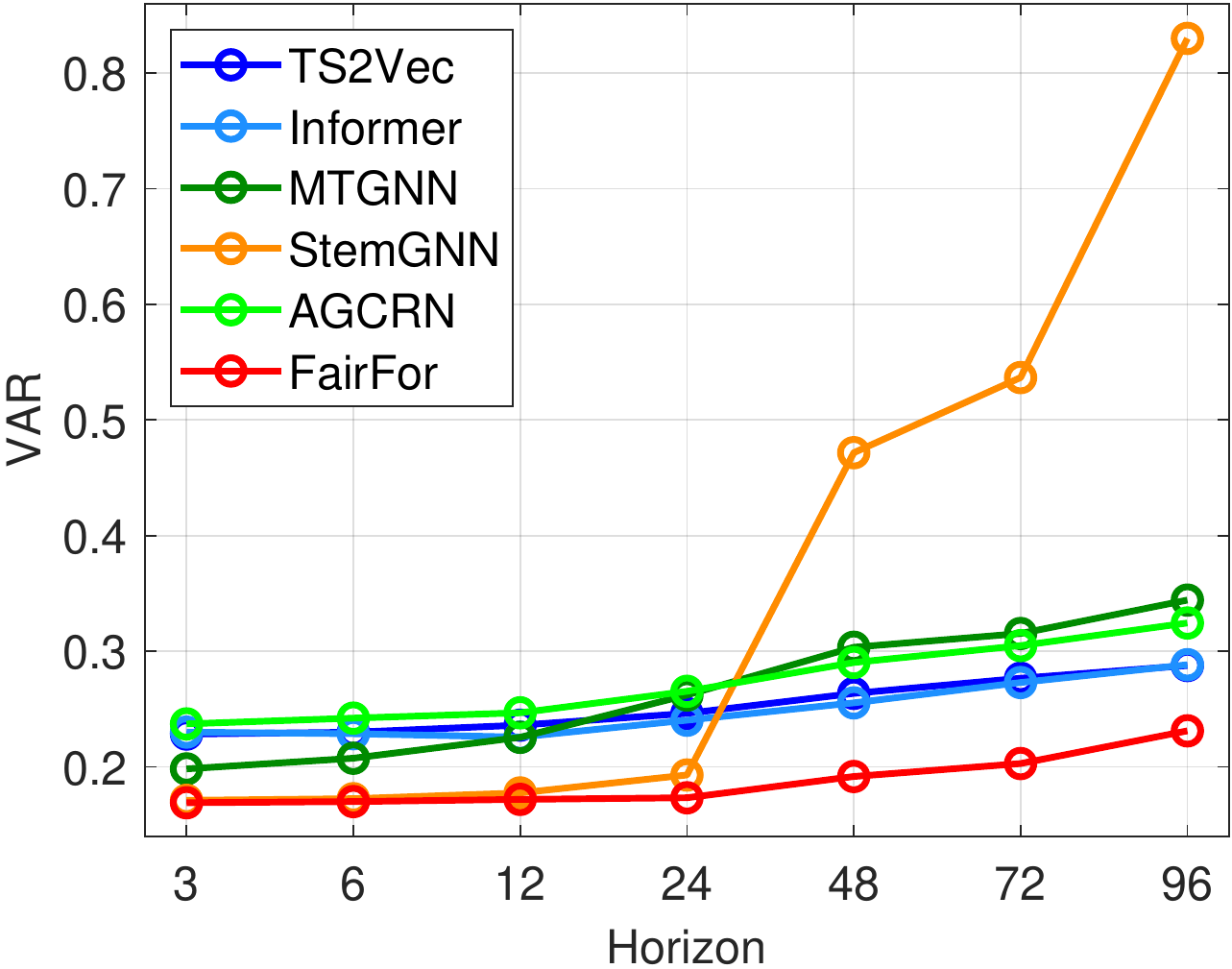}
        \vspace{0.3mm}
    \end{minipage}
    }
\centering
\caption{Performance comparison at different horizons.}
\label{figure3}
\end{figure*}

\begin{table*}[t]
 \newcommand{\tabincell}[2]{\begin{tabular}{@{}#1@{}}#2\end{tabular}}
 \centering
 \caption{Results of ablation study.}
 \label{table4}  
 \begin{tabular}{lc|c|c|c|c|c|c|c|c|c|c|c}
  \toprule 
  \multirow{2}{*}{Models } & \multicolumn{3}{c|}{PeMSD7(M)} & \multicolumn{3}{c|}{Solar-Energy} & \multicolumn{3}{c|}{Traffic} & \multicolumn{3}{c}{ECG5000} \\
  \cline{2-13}
  & MAE & MAPE & VAR  & MAE & MAPE & VAR  & MAE & MAPE & VAR & MAE & MAPE & VAR \\ 
  \midrule 
  w/o STCL+AGC & 4.2412 & 0.1018 & 30.2543 & 1.5797 & 3.4438 & 6.1499 & 0.0164 & 0.4758 & 5.93e-4 & 0.4500 & 0.9024 & 1.0522 \\
  w/o AL$\mathcal{D}$ & 2.8322 & 0.0682 & 20.4338 & 1.1407 & 3.3941 & 5.3096 & 0.0145 & 0.4666 & 4.16e-4 & 0.3168 & 0.9005 & 0.1752  \\
  w/o CL & 2.8571 & 0.0686 & 20.6593 & 1.1836 & 3.3942 & 5.1009 & 0.0112 & 0.3462 & 4.52e-4 & 0.3161 & 0.8974 & 0.1738  \\ 
  w/o ORL & 2.8377 & 0.0693 & 21.0896 & 1.2342 & 3.4104 & 4.8365 & 0.0122 & 0.3450 & 4.24e-4 & 0.3152 & 0.9005 & 0.1739  \\ 
  \midrule 
  \textbf{\textit{FairFor}} & \textbf{2.7170} & \textbf{0.0676} & \textbf{19.7870} & \textbf{1.0476} & \textbf{3.3539} & \textbf{4.7139} & \textbf{0.0103} & \textbf{0.2581} & \textbf{4.11e-4} & \textbf{0.3150} & \textbf{0.8931} & \textbf{0.1718}   \\ 
  \bottomrule 
 \end{tabular}
\end{table*}

\textbf{Forecasting Performance} From Table \ref{table3}, we  observe that \textit{FairFor} can still achieve high multi-step forecasting quality on PeMSD7(M) and ECG5000. \textit{FairFor} makes $3.54\%/4.05\%/2.63\%$ improvements in average w.r.t. MAE, RMSE and MAPE over the best baseline on PeMSD7(M) and ECG5000.
This is explainable: different from the motivation of sacrificing overall performance to guarantee the fairness requirement in existing recommendation methods~\cite{DBLP:conf/www/PatroBGGC20,DBLP:conf/www/WuCSHWW21}, \textit{FairFor} approaches an unbiased process of enriching the representation of per variable and thus can achieve superior performance of prediction and fairness simultaneously.
Although our model is surpassed on some specific examples, i.e. Solar-Energy and Traffic, the forecasting performance is still sub-optimal.
More specifically, we reveal the cause of \textit{FairFor}'s underperformance on Solar-Energy and Traffic datasets. Our method has a performance decrease over Informer of 8.42$\%$ (MAE), 9.30$\%$ (RMSE) and 3.99$\%$ (MAPE) on Traffic. This is intuitively attributable to the fact that the capture of mid-term (e.g., $h \in \{12, 24\}$) and long-term (e.g., $h \in \{48,72,96\}$) dependency  between output and input in Informer is more precise and contributes more to improve overall regression indicators, which is also verified in Figure \ref{Solar-Energy_MAE}, \ref{Traffic_MAE}. Moreover, MAE, RMSE and MAPE decrease of AGCRN over proposed \textit{FairFor} is 15.25$\%$, 0.49$\%$ and 1.93$\%$ on Solar-Energy respectively. This is attributable that fine-grained node-specific (e.g., similar, dissimilar, or contradictory) patterns across different data sources may be captured by node-specific parameter learning strategy instead of parameter sharing strategy commonly used in MTGNN, StemGNN and \textit{FairFor} (with GNN-based backbone). However, these methods neglect to focus on disadvantaged variables in design and cause poor VAR performance. 


Regarding computational expenses, we investigate the parameter volumes and time costs of \textit{FairFor} and two GNN-based competitive baselines AGCRN and StemGNN on PeMSD7(M). \textit{FairFor} has a parameter volume of 0.85M, while AGCRN and StemGNN have parameter volumes of 0.75M and 1.22M, respectively. Moreover, the training time per epoch for \textit{FairFor} is 32.83 seconds, which is lower than AGCRN (35.59 seconds) but higher than StemGNN (7.78 seconds). Despite the relatively large parameters and time costs associated with \textit{FairFor}, it has effectively addressed the challenging variable disparity issue and achieved a significant performance improvement over the baselines, see Table \ref{table2} and Table \ref{table3}. Therefore, the computation and time costs of \textit{FairFor} are considered moderate and acceptable.

Next, to verify the ability of forecasting different horizons, we compare \textit{FairFor} with TS2Vec, Informer, Pyraformer, MTGNN,  StemGNN and AGCRN, which are the top-6 baselines from the previous experiments. We fix $w=12$ and adjust $h$ in the range of $\{3,6,12,24,48,72,96\}$. Figure \ref{figure3} further shows the MAE and VAR performance comparison at different horizons. We can observe that: 1) the \textit{FairFor} shows significantly better VAR results at different horizons across all datasets; 2) On the Solar-Energy and Traffic datasets, Informer achieves better MAE results on the long horizon (at $\{48,72,96\}$). We attribute this to the potential value to capture the long-range dependencies.


\subsubsection{Ablation Study}
Further, we  perform an ablation study to  evaluate the effectiveness of several core designs in our \textit{FairFor}, i.e., spatio-temporal correlation learning enhanced by AGC (STCL+AGC) , adversarial learning - $\mathcal{D}$ (AL$\mathcal{D}$), clustering loss (CL), and orthogonal regularization loss (ORL), by removing them individually. 
We select the setting as $w=12,h=12,K=6$ on all datasets. 

Results are summarized in Table \ref{table4}. We  observe that: 
1) In \textbf{w/o STCL+AGC}, the STCL module enhanced by AGC module of \textit{FairFor} is replaced by gated TCN previously used in~\cite{DBLP:conf/ijcai/WuPLJZ19} to learn temporal dependencies. Table \ref{table4} shows our method averagely outperforms its variant \textbf{w/o STCL+AGC} with a significant margin (MAE: -30.14$\%$, MAPE: -16.04$\%$, VAR: -40.55$\%$) across all datasets, indicating that jointly capturing the temporal patterns and inter-series correlations can significantly improve the forecasting and fairness performance.
2) Variant \textbf{w/o AL$\bm{\mathcal{D}}$} denotes \textit{FairFor} without discriminator $\mathcal{D}$, and $\mathcal{L}_{A}$ is removed from training objective $\mathcal{L}$ in Eq. (\ref{eqn16}). As verified in Table \ref{table4}, the scores of MAE, MAPE, and VAR increases by 10.44$\%$, 11.89$\%$, and 4.38$\%$ on average. This proves that group-based adversarial learning is worth adopting, especially when balancing and optimizing the performance of each variable is required. 
3) To study the importance of the variable correlating $\&$ grouping module without affecting the normal work of discriminator $\mathcal{D}$, we only discard $\mathcal{L}_C$ from $\mathcal{L}$ in Eq. (\ref{eqn16}) to form variant \textbf{w/o CL}. It yields a 6.20$\%$/7.15$\%$/5.51$\%$ rise on average across all datasets, demonstrating the significance of setting a soft K-means objective to auxiliarily infer variable correlations and group variables in group-based adversarial training.
4) Variant \textbf{w/o ORL} represents discarding $\mathcal{L}_D$ from $\mathcal{L}$ in the training process. The results demonstrate that removing ORL also hurts the fairness of \textit{FairFor}. Because ORL promotes the orthogonality of group-relevant and -independent representation, which can better identify and remove group-relevant information. Note that \textit{FairFor} achieves relatively slight improvement over variants \textbf{w/o ALD}, \textbf{w/o CL}, and \textbf{w/o ORL} on the ECG5000 dataset. This may be attributed to the fact that the average inter-series distance (11.15) after normalization on the ECG5000 dataset is significantly smaller than that on other datasets, e.g., the Solar-Energy and PeMSD7(M) datasets which have an average inter-series distance of 29.45 and 41.28, respectively. As a result, the grouping/clustering-related modules (i.e., ALD, CL, and ORL) may be less sensitive to inter-series distances and therefore perform less effectively in grouping/clustering time series on the ECG5000 dataset.

The STCL module enhanced by AGC exhibits the greatest impact on the performance because it has resorted to distilling complex temporal patterns and inter-series correlations as the basis of time-series representation. Besides, other modules enhance the spatio-temporal representation learning from the perspective of learning more informative representations attending to both advantaged and disadvantaged variables. The ablation variants except for \textbf{w/o STCL+AGC} still outperform most SOTA baselines, e.g., the MAE/MAPE/VAR scores of \textbf{w/o ORL} are average 9.10$\%$/11.17$\%$/18.46$\%$ lower than StemGNN across all datasets and are average 4.70$\%$/4.38$\%$/18.35$\%$ lower than AGCRN on PeMSD7(M) and ECG5000, indicating the stable and complementary advantages of different modules.

\begin{figure}[!t]
\centering
\subfigure[Number of Clusters $K$]{
    \label{K}
    \includegraphics[width=0.47\linewidth]{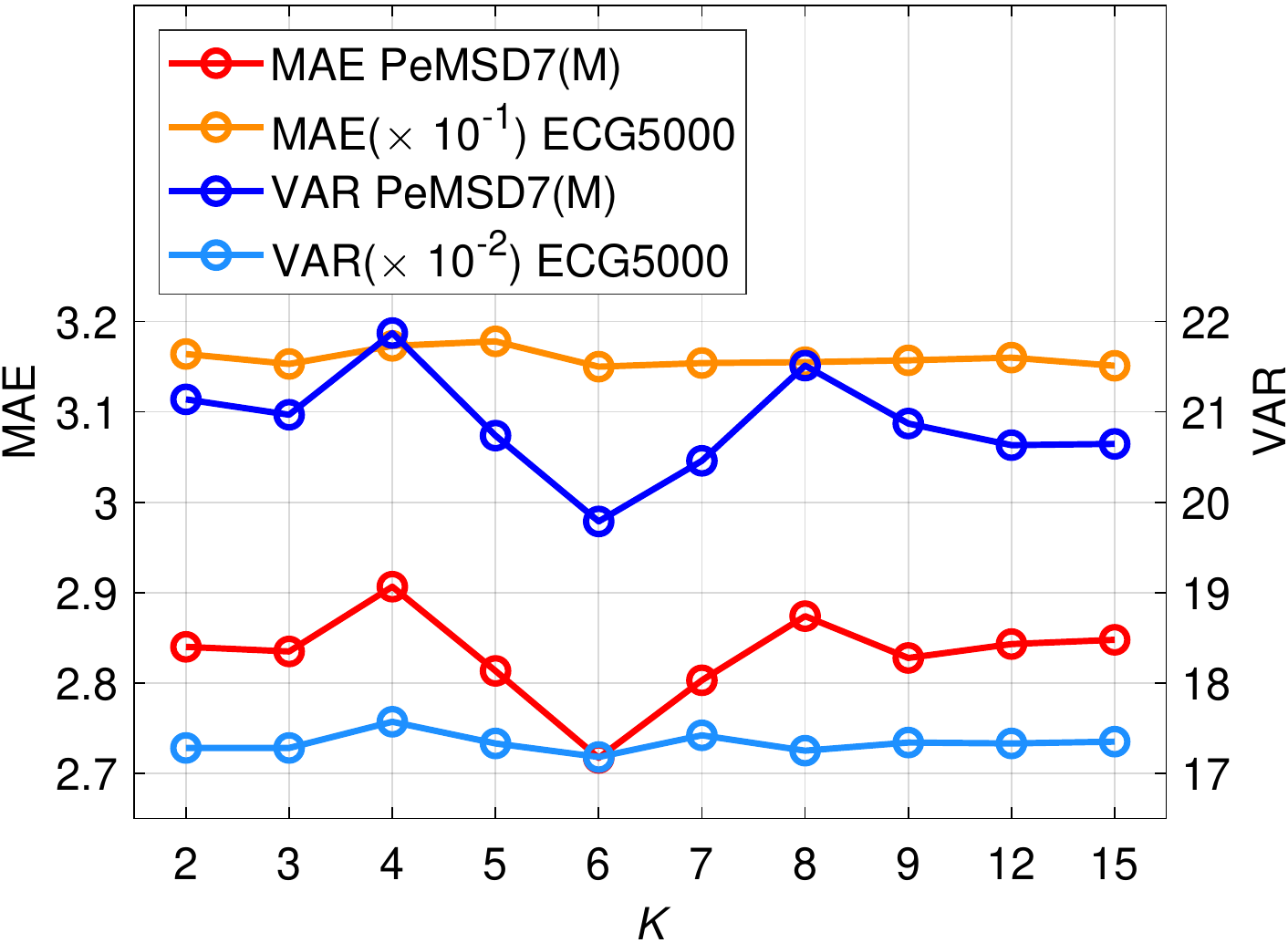 }
    }
\subfigure[Window Size $w$]{
    \label{w}
    \includegraphics[width=0.47\linewidth]{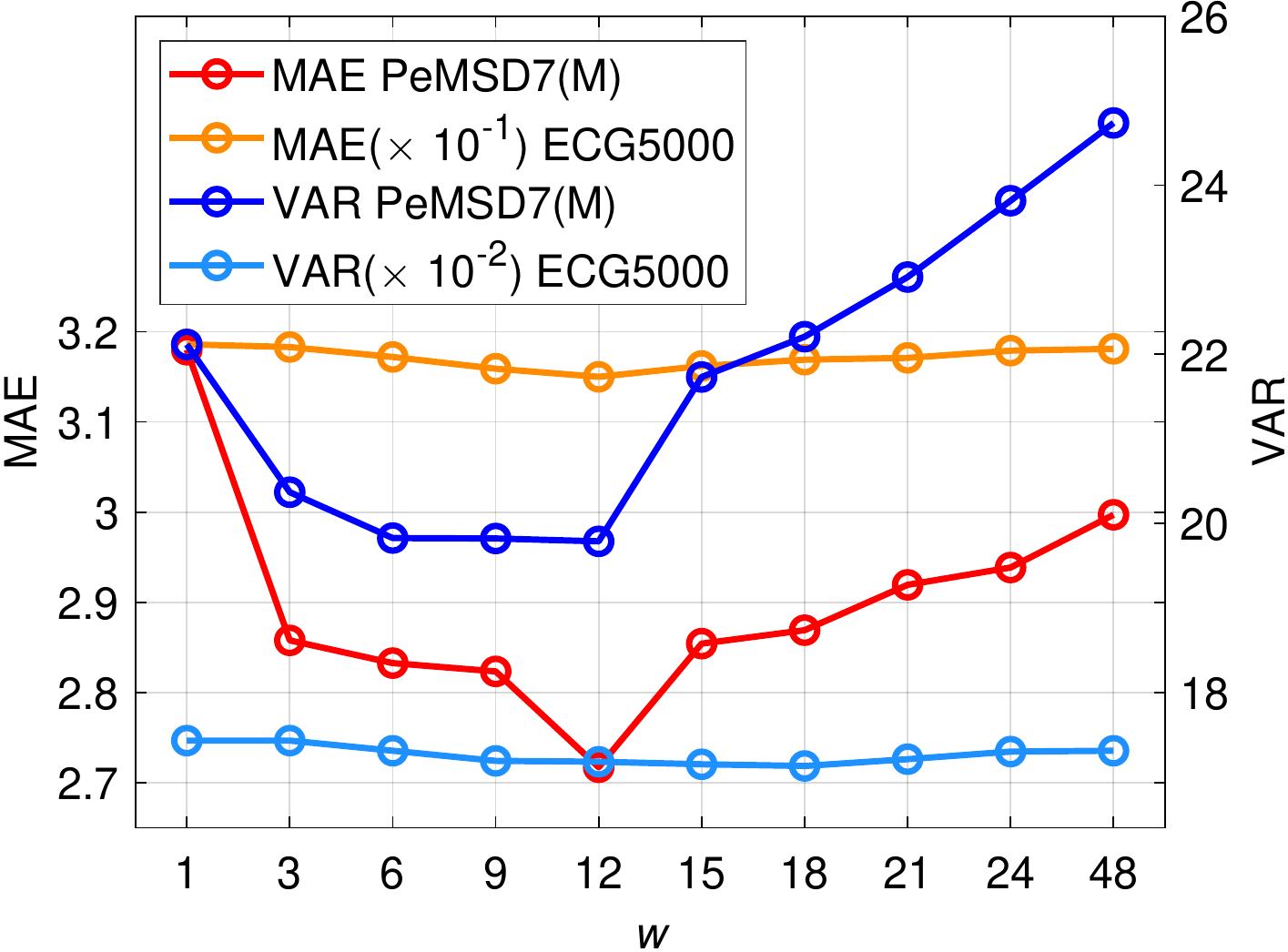}
    }
\subfigure[Embedding Dimension $d$]{
    \label{d}
    \includegraphics[width=0.47\linewidth]{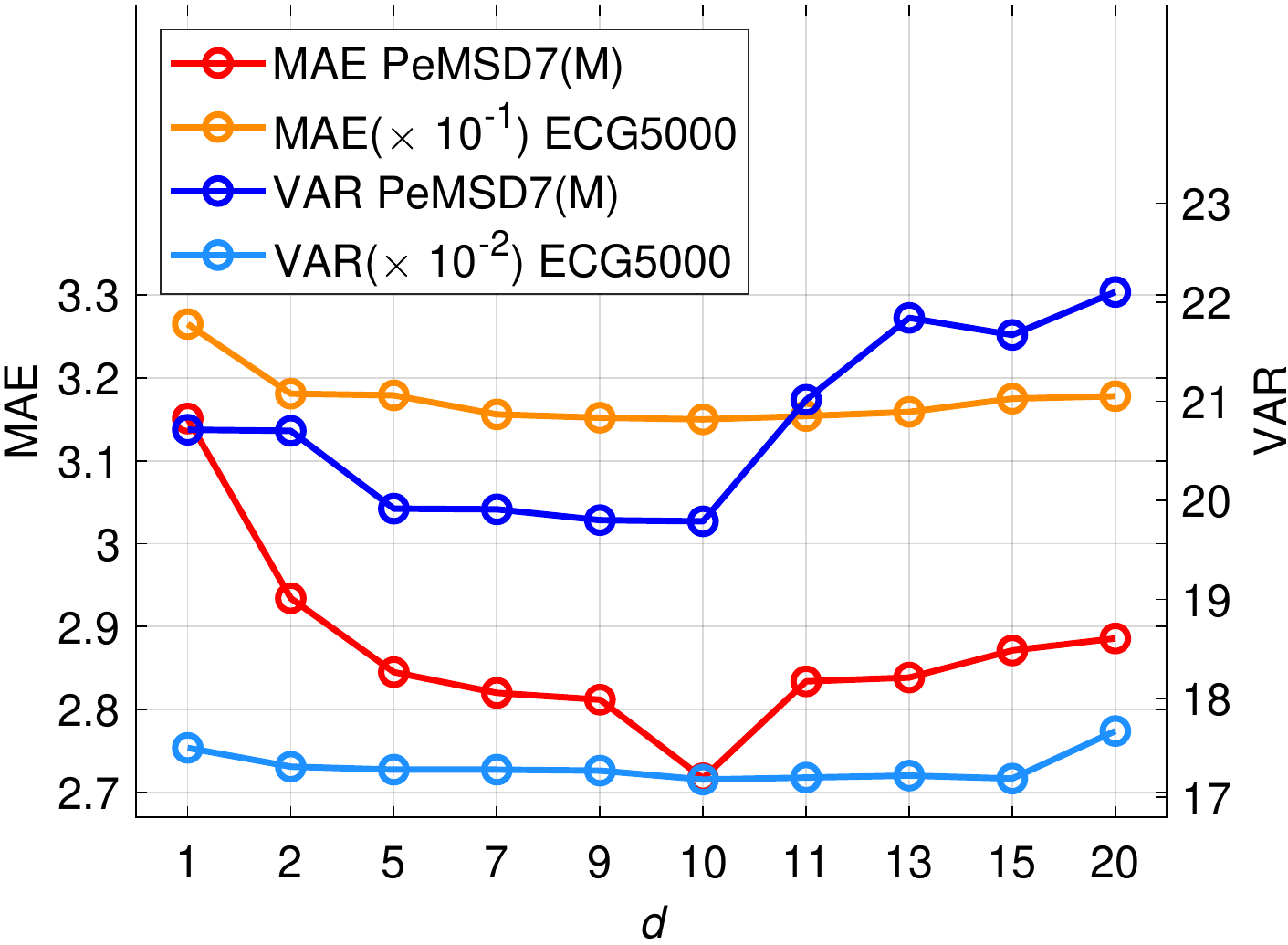}
    }
\subfigure[Trade-off Coefficient $\lambda_{a}$]{
    \label{lambda1}
    \includegraphics[width=0.47\linewidth]{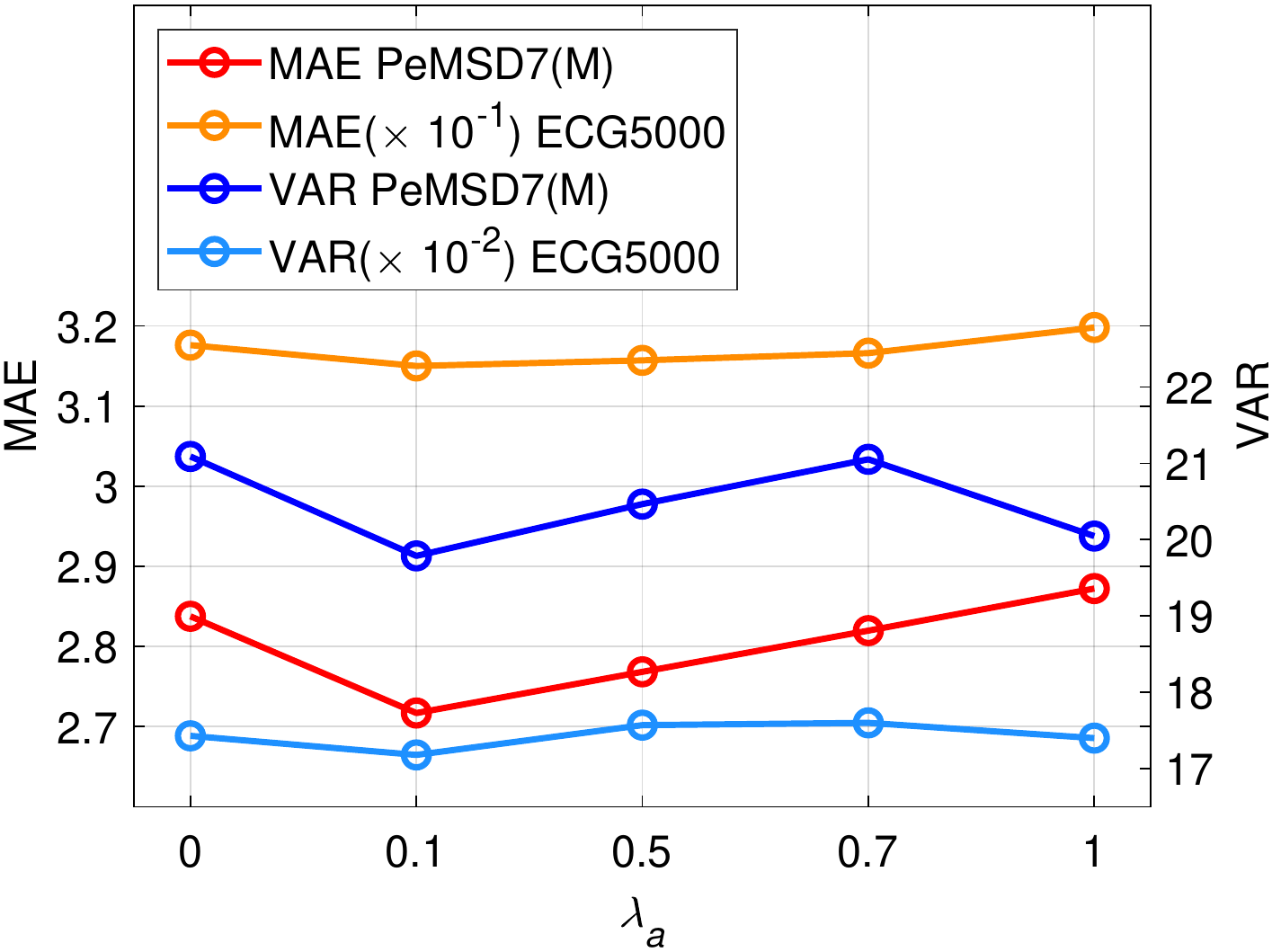}
    }
\subfigure[Number of Neighbors $n$ on PeMSD7(M)]{
    \label{n_1}
    \includegraphics[width=0.47\linewidth]{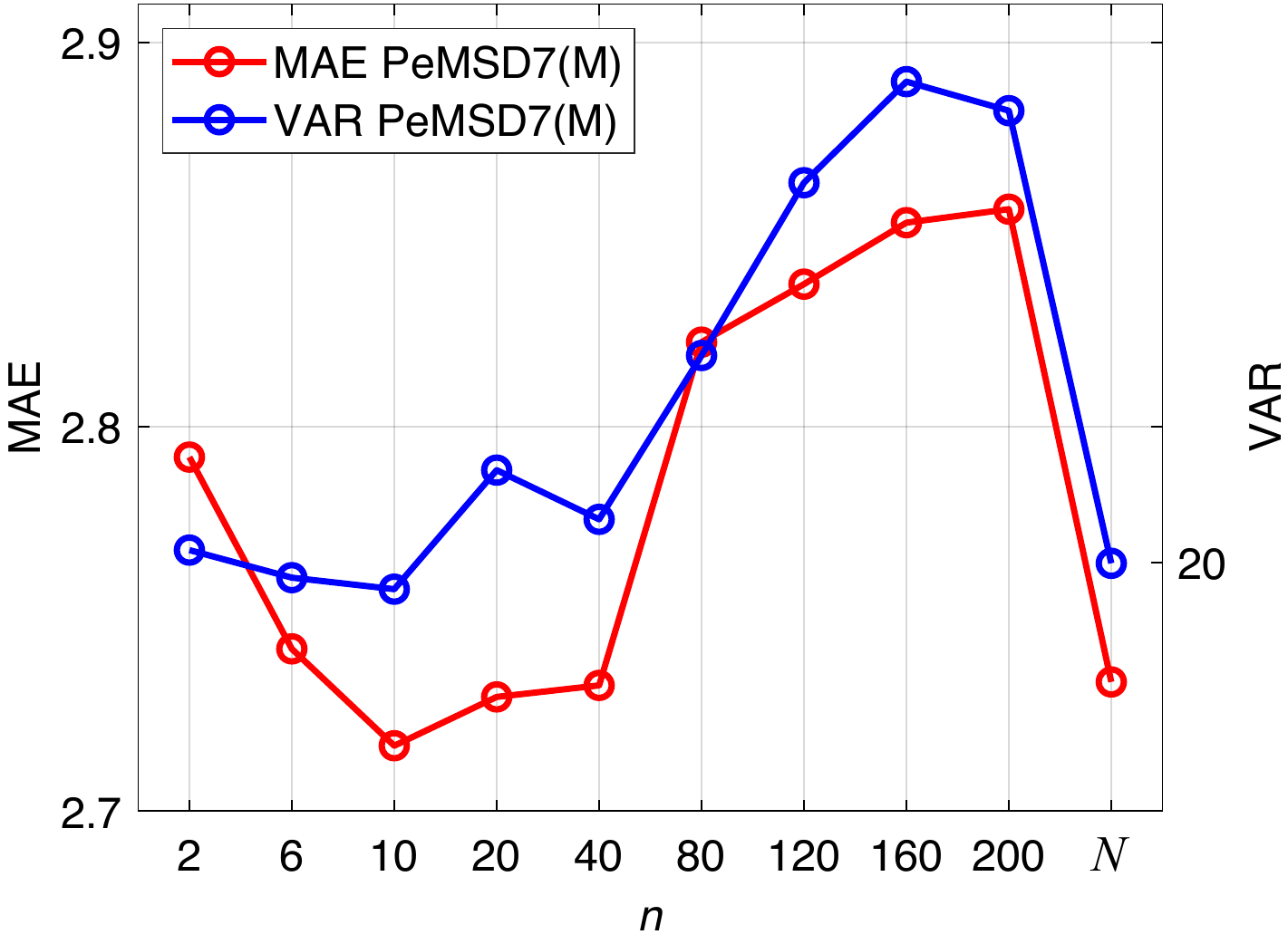}
    }
\subfigure[Number of Neighbors $n$ on ECG5000]{
    \label{n_2}
    \includegraphics[width=0.47\linewidth]{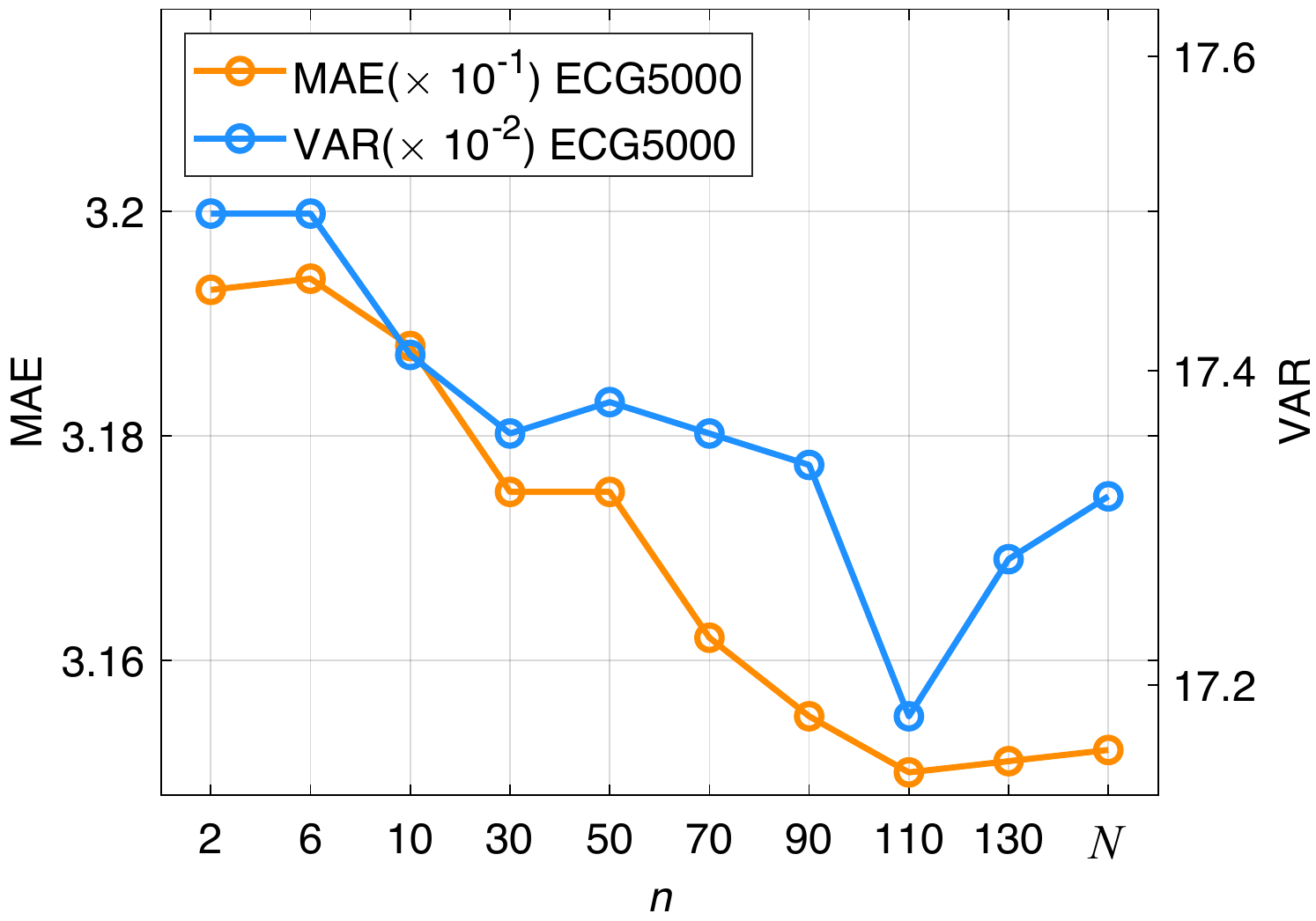}
    }
\centering
\caption{Hyper-parameter analysis on PeMSD7(M) and ECG5000 datasets.}
\label{figure4}
\end{figure}

\begin{figure*}[!t]
\centering
\subfigure[$\bm{H}$ on Traffic]{
    \label{h-traffic}
    \includegraphics[width=0.22\linewidth]{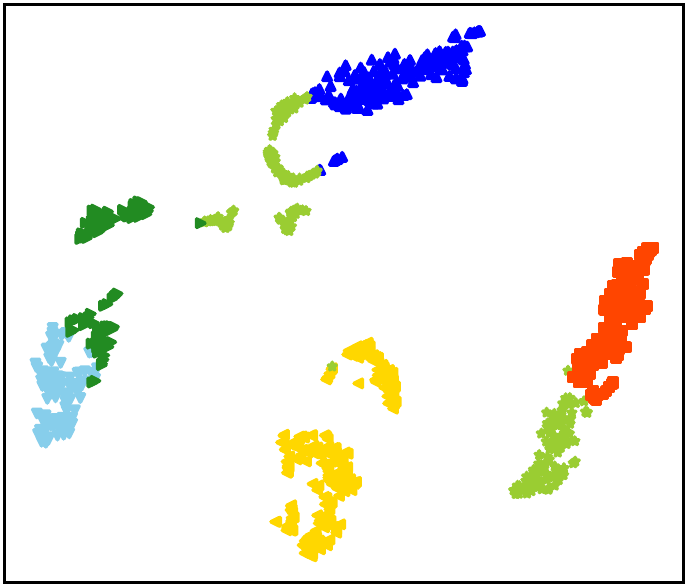}
    }   
\subfigure[$\bm{H}$ on PeMSD7(M)]{
    \label{h-pems}
    \includegraphics[width=0.22\linewidth]{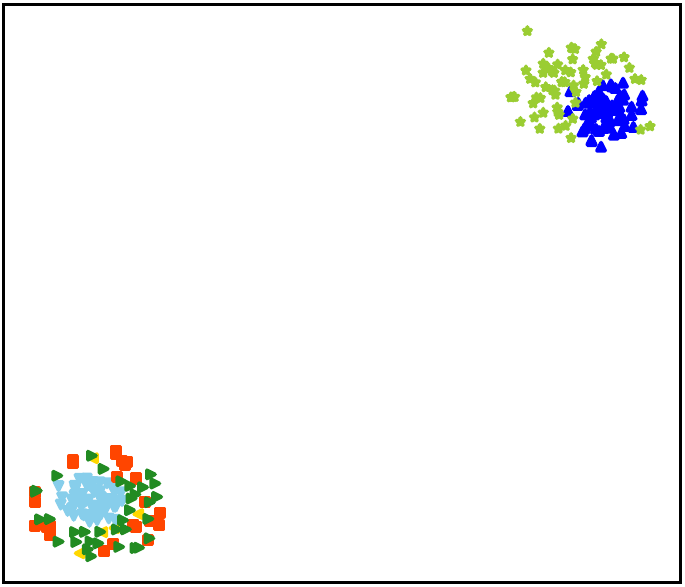}
    }
\subfigure[$\bm{H}$ on Solar-Energy]{
    \label{h-solar}
    \includegraphics[width=0.22\linewidth]{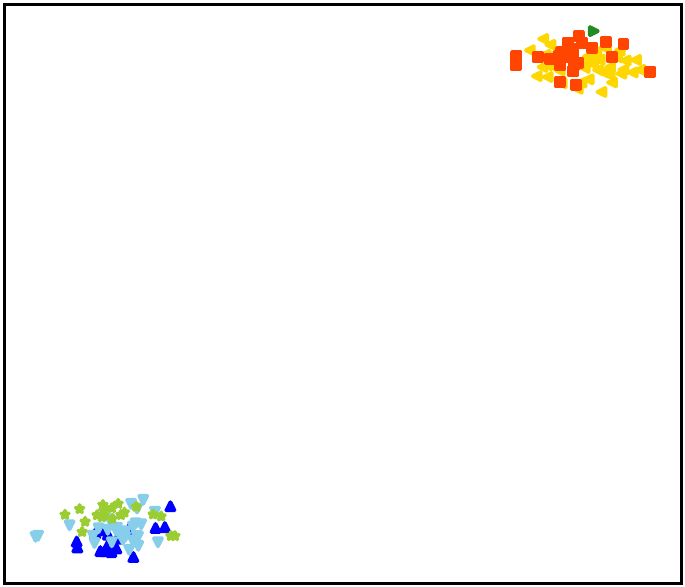}
    }
\subfigure[$\bm{H}$ on ECG5000]{
    \label{h-ECG5000}
    \includegraphics[width=0.22\linewidth]{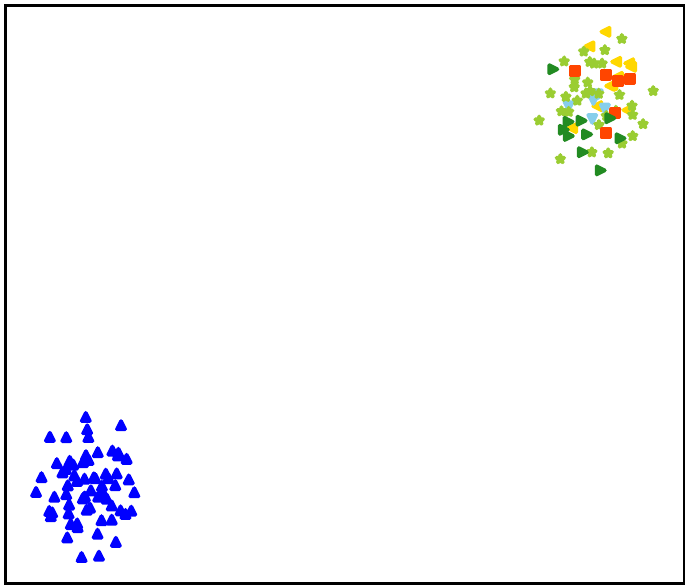}
    }
\subfigure[$\widehat{\bm{H}}$ on Traffic]{
    \label{h-hat-traffic}
    \includegraphics[width=0.22\linewidth]{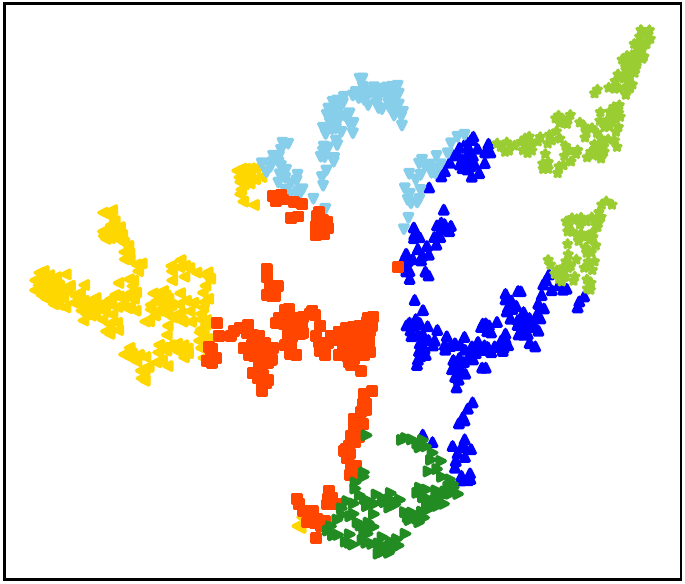}
    }%
\hspace{0.03cm}
\subfigure[$\widehat{\bm{H}}$ on PeMSD7(M)]{
    \label{h-hat-pems}
    \includegraphics[width=0.22\linewidth]{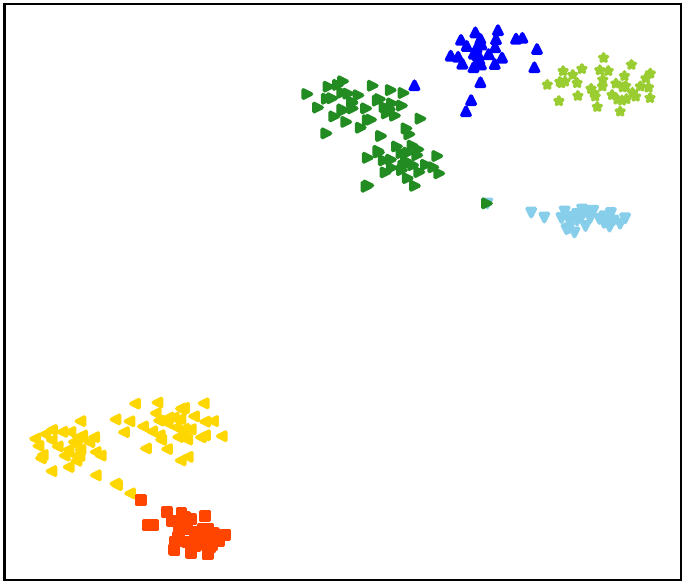}
    }
\subfigure[$\widehat{\bm{H}}$ on Solar-Energy]{
    \label{h-hat-solar}
    \includegraphics[width=0.22\linewidth]{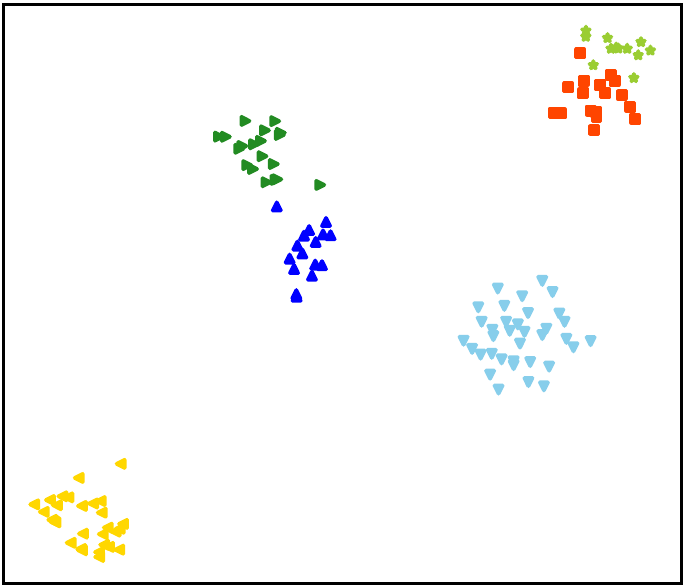}
    }
\subfigure[$\widehat{\bm{H}}$ on ECG5000]{
    \label{h-hat-ECG5000}
    \includegraphics[width=0.22\linewidth]{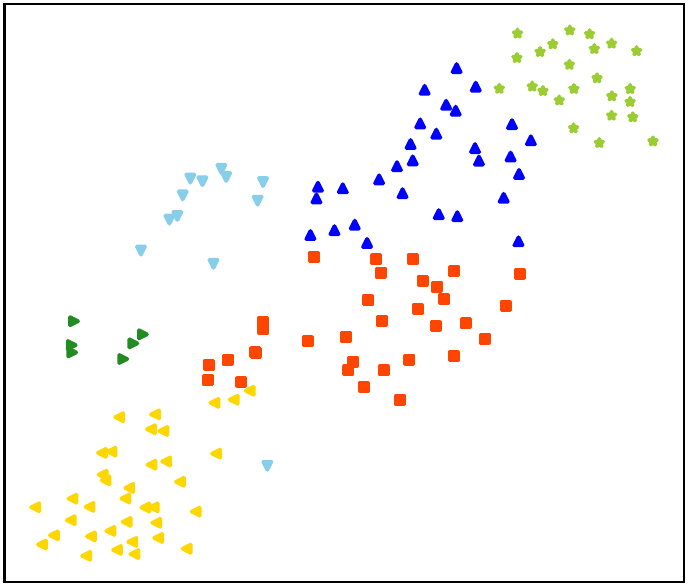}
    }
\centering
\caption{$t$-SNE feature distribution visualization of $\bm{H}$ and $\widehat{\bm{H}}$.}
\label{figure5}
\end{figure*}

\subsubsection{Hyper-parameter Analysis}
We explore the sensitivity analysis of several critical hyper-parameters on PeMSD7(M) and ECG5000 datasets, including the number of clusters $K$, window size $w$, embedding dimension $d$, trade-off coefficient $\lambda_{a}$ and the number of neighbors $n$. The results in terms of MAE and VAR are demonstrated in Figure \ref{figure4}. When we change a hyper-parameter, the other variables keep their default values explained in Section 4.2.

First, we investigate the influence of variable correlating $\&$ grouping by varying the value of $K$ in a range of $\{2,3,4,5,6,7,8,9,12,15\}$, and plot the forecasting results MAE and fairness results VAR under $w=12,h=12$ in Figure \ref{K}. When $K$ is less than 6, MAE and VAR display a trend of violent fluctuation. A moderate value for $K$, i.e., $K=6$, makes MAE and VAR descend to the lowest. A similar trend exists as increasing cluster number from 6 since unsuitable clustering objective possibly cannot fully describe the complex group information inside variables and confuse latent data regularities~\cite{DBLP:journals/corr/abs-1810-07758}. Meanwhile, the increase of $K$ will bring a burden to the filtering $\&$ fusion module and hurt the performance.
Note that the performance on ECG5000 is relatively stable with different $K$. The possible reason is that the inter-series distance on the ECG5000 dataset is significantly smaller than that on other datasets, which makes the model less sensitive to variations in $K$. 

Then, we turn to explore the impact of input window with different sizes by varying $w$ in a range of $\{1,3,6,9,12,15,18,21,24,48\}$. As shown in Figure \ref{w}, when predicting mid-term sequence ($h=12$), initially increasing the window size causes the MAE and VAR to drop since it brings sufficient temporal patterns and dependency information, but further increasing $w$ makes  MAE and VAR  higher. This is intuitive due to the noise introduced by redundant temporal patterns, dependencies and group features that do not fit the group-based adversarial learning task. Hence, the input window size and output window size need to be carefully weighed.

Moreover, we check a key parameter $d$ of learnable node embedding matrix $\bm{E}_{\bm{M}}$, which not only influences the quality of adjacent matrix $\widehat{\bm{M}}$ but also decides the parameter diversity in the STCL module (Eq. (\ref{eqn6}), (\ref{eqn7}) and (\ref{eqn8})). Figure \ref{d} reports the effects taking different dimensions of $\bm{E}_{\bm{M}}$. As can be seen, \textit{FairFor} achieves relatively good performance with all tested dimensions, demonstrating that the model capacity is fairly robust. Note that \textit{FairFor} achieves the best performance when the dimension is set to 10. As the dimension goes larger, the performance first increases since more information is contained which thus helps our STCL module to deduce more accurate intra-series correlations, but getting weaker probably due to over-fitting. Intuitively, an appropriate embedding dimension should make the model obtain sufficient correlation information meanwhile avoiding introducing additional parameters causing over-fitting.

To learn the importance of adversarial losses, we control $\lambda_{a}$ to change within a range $\{0,0.1,0.5,0.7,1\}$. The results are exhibited in Figure \ref{lambda1}. From Figure \ref{lambda1}, we find that the MAE drops steadily with the increase of $\lambda_{a}$, then rises until $\lambda_{a}=1$, and the VAR has a similar trend. When $\lambda_{a}=0.1$, forecasting performance and fairness achieve the best together. Thus, a moderate value for $\lambda_{a}$ (e.g., 0.1) may be preferable to make adversaries achieve an appropriate equilibrium and protect group information from being leaked to the group-irrelevant representation.

Finally, we probe into investigating different numbers of neighbors $n$ from 2 to $N$ for the graph sparse strategy, the results on PeMSD7(M) and ECG5000 are reported in Figure \ref{n_1} and Figure \ref{n_2} respectively. We observe here that selecting the top-$10$ and top-$110$ closest nodes (the best number of neighbors, denoted as $n^{\star}$) as neighbors yields the maximum benefit to the prediction task for PeMSD7(M) and ECG5000 respectively. When $n$ increases from 2 to $n^{\star}$, the MAE and VAR decrease gradually as expected, revealing the merit of selecting sufficient neighbors to disclose the underlying inter-series dependencies. Furthermore, when $n^{\star} \textless n \leq N$, the MAE and VAR increase. This is intuitive because only a few inter-series/node connections are meaningful enough for each node and increasing its neighborhood merely introduces more additional noise.

\subsubsection{Visualization of Fairness}
As previously discussed, the filtering $\&$ fusion module filters the group-relevant information from $\bm{H}$ and obtains the group-irrelevant representation $\widehat{\bm{H}}$ in the group-based adversarial training. To further understand how the group-relevant (specific to each group) and -independent (shared by all groups) representation works, we respectively visualize the feature distribution by $t$-SNE in 2-D space and evaluate the informativeness of $\bm{H}$ and $\widehat{\bm{H}}$.
We randomly choose a batch of the Traffic-train data. 
Figure \ref{h-traffic}-\ref{h-ECG5000} and Figure \ref{h-hat-traffic}-\ref{h-hat-ECG5000} show the feature distributions of $\bm{H}$ and $\widehat{\bm{H}}$, respectively. Firstly, both $\bm{H}$ and $\widehat{\bm{H}}$  exhibit a good clustering structure, which proves the clustering ability of our model. In addition, $\bm{H}$ shows smaller inter-cluster distances than $\widehat{\bm{H}}$, indicating different variable representations learned in $\bm{H}$ are close and concentrated, while those learned in $\widehat{\bm{H}}$ after filtering are informative and discriminative and exhibit less overlap. Furthermore, $\bm{H}$ with a compact clustering structure may only contain the information of original advantaged groups, and $\widehat{\bm{H}}$ with a decentralized clustering structure is enriched by drawing support from advantaged groups and changes into informative representation shared by all groups.
Therefore, the \textit{FairFor} model with good prediction and fairness performance is also interpretable.

\section{Conclusion}
We argue that there could be performance unfairness across MTS variables due to the variable disparity, hence equally attending to both advantaged and disadvantaged variables may facilitate overall forecasting performance and fairness simultaneously. In this paper, we study the unfairness problem and propose a fairness-aware MTS forecasting method (\textit{FairFor}). FairFor aims to (1) capture spatio-temporal variable correlations via recurrent graph convolution, (2) group variables by leveraging a spectral relaxation of the K-means objective, (3) filter the group-relevant information and generate group-irrelevant representation via adversarial learning with an orthogonality regularization, and (4) integrate the group-relevant and -irrelevant representations to form a highly informative representation for the final prediction. 
Experimental results on four real-world datasets prove that \textit{FairFor} can simultaneously and effectively improve the performance of fairness and MTS forecasting. Currently, \textit{FairFor} is an enhanced framework that sequentially performs to capture spatio-temporal correlations and balance the forecasting performance of all variables. Intuitively, the variable disparity and correlations dynamically evolve over time in coordination. Therefore, we will embed the balance operation into spatio-temporal correlation learning at each time step. Additionally, we will apply this work to more real-world scenarios such as enterprise innovation analysis in the future work.

\ifCLASSOPTIONcompsoc
  \section*{Acknowledgments}
\else
  \section*{Acknowledgment}
\fi

This work is supported by the National Natural Science Foundation of China under Grant 62272048.

\ifCLASSOPTIONcaptionsoff
  \newpage
\fi



\bibliographystyle{IEEEtran}
\bibliography{ref}
%



%





\begin{IEEEbiography}[{\includegraphics[width=1in,height=1.25in,clip,keepaspectratio]{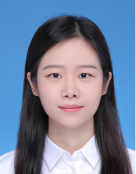}}]{Hui He} received the ME degree from University of Shanghai for Science and Technology, Shanghai, China in 2020. She is currently working toward the PhD degree with the School of Medical Technology, Beijing Institute of Technology, Beijing, China. Her current research interests focus on multivariate time-series analysis and knowledge services.
\end{IEEEbiography}

\begin{IEEEbiography}[{\includegraphics[width=1in,height=1.25in,clip,keepaspectratio]{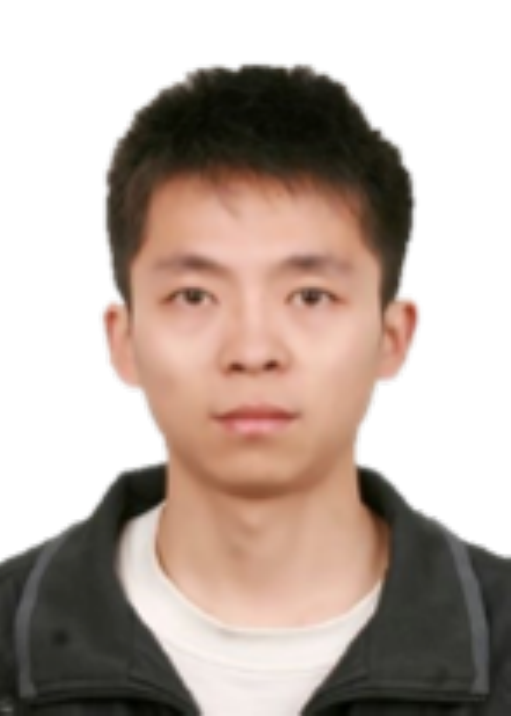}}]{Qi Zhang}
received the PhD degree from the Beijing Institute of Technology, Beijing, China and the University of Technology Sydney, Sydney, NSW, Australia, in 2020, under the dual PhD program. 
He is currently a Research Fellow with Tongji University, Shanghai, China. He has authored high-quality papers in premier conferences and journals, including AAAI Conference on Artificial Intelligence (AAAI), International Joint Conferences on Artificial Intelligence (IJCAI), 
International World Wide Web Conference (TheWebConf), IEEE Transactions on Knowledge and Data Engineering (TKDE)
and ACM Transactions on Information Systems (TOIS). His primary research interests include collaborative filtering, sequential recommendation, learning to hash, and MTS analysis.
\end{IEEEbiography}

\begin{IEEEbiography}[{\includegraphics[width=1in,height=1.25in,clip,keepaspectratio]{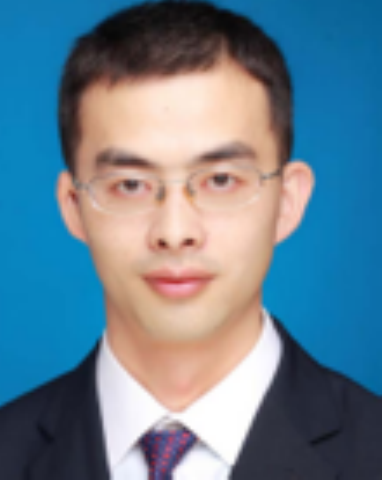}}]{Shoujin Wang}
received the PhD degree in data science from the University of Technology Sydney (UTS), Sydney, NSW, Australia, in 2019. 
He is currently a Lecturer in data science with UTS. He has authored high-quality papers in premier conferences and journals, including International World Wide Web Conference (TheWebConf), AAAI Conference on Artificial Intelligence (AAAI), International Joint Conferences on Artificial Intelligence (IJCAI), 
and ACM Computing Surveys (ACMCSUR). His research interests include data mining, machine learning, recommender systems, and fake news mitigation.
He was a recipient of some prestigious awards, including the 2022 DSAA Next-generation Data Scientist Award and the 2022 Club Melbourne Fellowship Award. 
\end{IEEEbiography}

\begin{IEEEbiography}[{\includegraphics[width=1in,height=1.25in,clip,keepaspectratio]{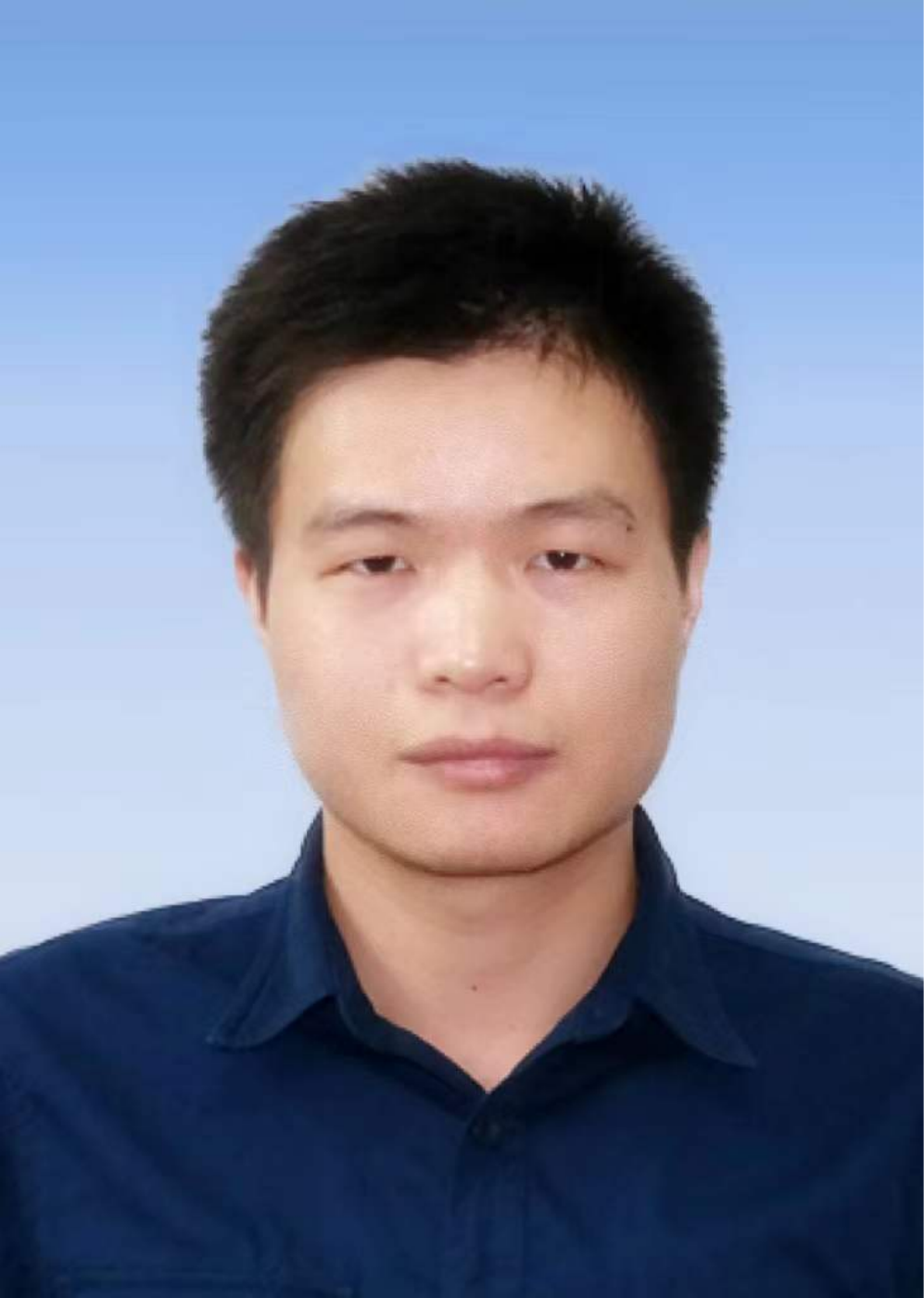}}]{Kun Yi}
is a currently working toward the PhD degree with the Beijing Institute of Technology, China. His current research interests include multivariate time-series forecasting, data science and knowledge discovery.
\end{IEEEbiography}

\begin{IEEEbiography}[{\includegraphics[width=1in,height=1.25in,clip,keepaspectratio]{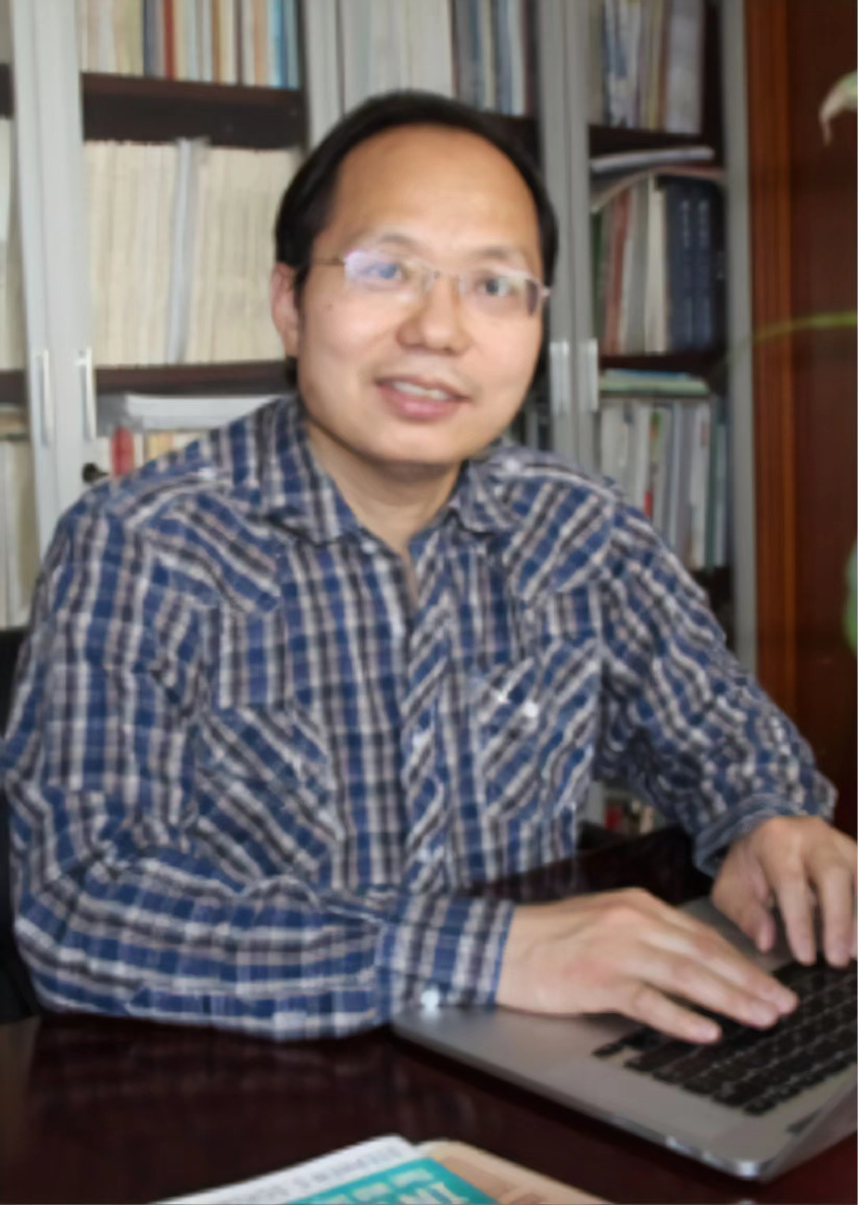}}]{Zhendong Niu}
received the PhD degree in computer science from the Beijing Institute of Technology in 1995. 
He was a post-doctoral researcher with the University of Pittsburgh, Pittsburgh, PA, USA, from 1996 to 1998, a researcher/adjunct faculty member with Carnegie Mellon University, Pittsburgh, from 1999 to 2004, and a joint research professor with the School of Computing and Information, University of Pittsburgh, in 2006. 
He is a professor with the School of Computer Science and Technology, Beijing Institute of Technology, Beijing. His research interests include informational retrieval, software architecture, digital libraries, and web-based learning techniques. 
He received the International Business Machines Corporation (IBM) Faculty Innovation Award in 2005 and the New Century Excellent Talents in University of the Ministry of Education of China in 2006.  
\end{IEEEbiography}

\begin{IEEEbiography}[{\includegraphics[width=1in,height=1.25in,clip,keepaspectratio]{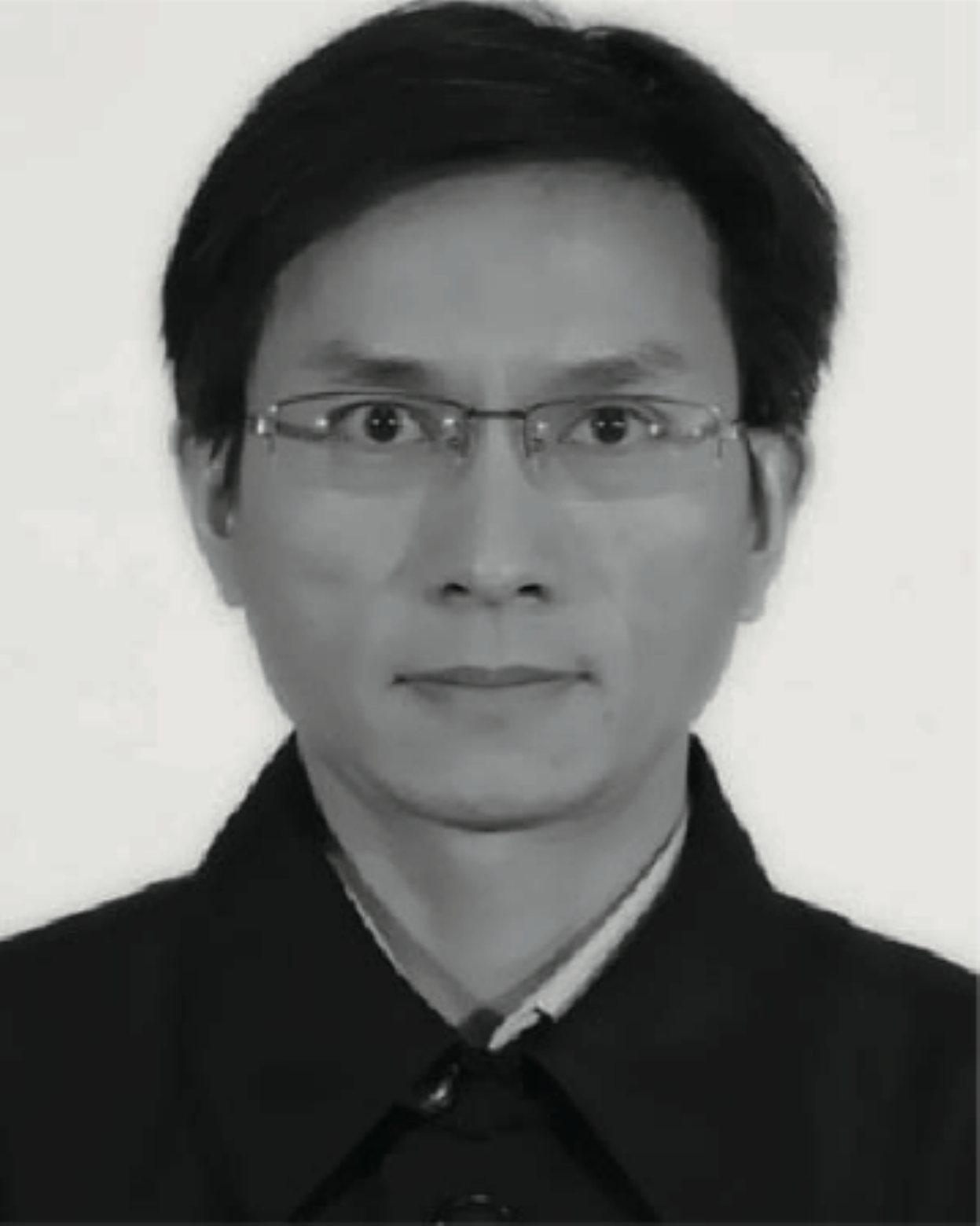}}]{Longbing Cao} (@SM in 2006) received PhD degree in pattern recognition and intelligent systems from the Chinese Academy of Science, China, and the PhD degree in computing sciences from the University of Technology Sydney, Australia. He is a distinguished chair professor with Macquarie University, an ARC future fellow (professorial level), and the EiCs of IEEE Intelligent Systems and J. Data Science and Analytics. His research interests include artificial intelligence, data science, machine learning, behavior informatics, and their enterprise applications.
\end{IEEEbiography}




\end{document}